\documentclass[lettersize,journal]{IEEEtran}
\usepackage{amsmath,amsfonts}
\usepackage{algorithmic}
\usepackage{algorithm}
\usepackage{array}
\usepackage[caption=false,font=normalsize,labelfont=sf,textfont=sf]{subfig}
\usepackage{textcomp}
\usepackage{stfloats}
\usepackage{url}
\usepackage{verbatim}
\usepackage{graphicx}

\usepackage{multicol}
\usepackage{booktabs} 
\usepackage{amssymb}  
\usepackage{multirow} 
\usepackage{amssymb}
\usepackage{natbib}
\usepackage{xcolor}

\begin{document}

\title{CA-YOLO: Cross Attention Empowered YOLO for \\Biomimetic Localization}

\author{~\IEEEmembership{Zhen Zhang\textsuperscript{*1},Qing Zhao\textsuperscript{*1,2},Xiuhe Li\textsuperscript{2},Cheng Wang\textsuperscript{2},Guoqiang Zhu\textsuperscript{2},Yu Zhang\textsuperscript{2},\\Yining Huo\textsuperscript{2},Hongyi Yu\textsuperscript{2},Yi Zhang\textsuperscript{\#2}}

\thanks{\textsuperscript{*} These authors made equal contributions to the research.}
\thanks{\textsuperscript{\#} correspondence: zy\_zhangyi@nudt.edu.cn}
\thanks{\textsuperscript{1} School of Mechanical and Automotive Engineering, Anhui Polytechnic University, Wu’hu 241000, Anhui, China.}
\thanks{\textsuperscript{2} College of Electrical and Electronic Engineering, National University of Defense Technology, He’fei 230000, Anhui, China.}
}

\markboth{Journal of \LaTeX\ Class Files,~Vol.~14, No.~8, August~2021}%
{Shell \MakeLowercase{\textit{et al.}}: A Sample Article Using IEEEtran.cls for IEEE Journals}


\maketitle

\begin{abstract}
In modern complex environments, achieving accurate and efficient target localization is essential in numerous fields. However, existing systems often face limitations in both accuracy and the ability to recognize small targets. In this study, we propose a bionic stabilized localization system based on CA-YOLO, designed to enhance both target localization accuracy and small target recognition capabilities. Acting as the "brain" of the system, the target detection algorithm emulates the visual focusing mechanism of animals by integrating bionic modules into the YOLO backbone network. These modules include the introduction of a small target detection head and the development of a Characteristic Fusion Attention Mechanism (CFAM). Furthermore, drawing inspiration from the human Vestibulo-Ocular Reflex (VOR), a bionic pan-tilt tracking control strategy is developed, which incorporates central positioning, stability optimization, adaptive control coefficient adjustment, and an intelligent recapture function. 
The experimental results show that CA-YOLO outperforms the original model on standard datasets (COCO and VisDrone), with average accuracy metrics improved by 3.94\%and 4.90\%, respectively.
Further time-sensitive target localization experiments validate the effectiveness and practicality of this bionic stabilized localization system. 
\end{abstract}

\section{Introduction}
\IEEEPARstart
{E}{xisting} 
target detection and localization systems often face challenges in stability and accuracy when dealing with small targets, variable-speed moving targets, and complex backgrounds \cite{ref1}. The main problems are as follows: small targets are difficult to accurately recognize due to weak features and few pixels; variable-speed moving targets may lead to unstable detection frames and missed detections; and lighting changes and viewing angle differences in complex backgrounds further increase the difficulty of target recognition. In addition, traditional localization systems often remain passive when the target is about to move out of the field of view, and cannot effectively respond to target's movement, lead to target loss and inability to continue tracking, which limit the performance and reliability of existing systems.

To overcome the challenges of small target recognition and the instability of detecting variable-speed targets, this study proposes a bionic stable positioning system based on CA-YOLO. By emulating the visual focusing mechanism found in animals, this system optimizes the target detection algorithm and introduces an efficient CA-YOLO network. Furthermore, a pan-tilt positioning system, designed in accordance with bionic principles, is developed to achieve both accurate small target detection and stable tracking of targets with variable speeds.

The CA-YOLO model incorporates three critical enhancements over existing approaches: A multi-head self-attention (MHSA) mechanism is integrated to improve the model’s ability to capture fine details; A specialized small target detection head is designed to enhance recognition accuracy for small targets \cite{ref2}; A channel and spatial attention module is introduced to optimize feature fusion. These innovations enable CA-YOLO to achieve high accuracy and maintain real-time performance, even in challenging and dynamic environments.

Drawing inspiration from the visual stability mechanisms observed in nature, particularly the human vestibulo-ocular reflex (VOR), this study designs a bionic pan-tilt positioning system. By simulating biological visual perception and motion control mechanisms, the system enables precise multi-degree-of-freedom control. Even in the presence of destabilizing factors such as vibration or movement, the system is capable of maintaining stable target tracking, thereby significantly enhancing both its stability and robustness.

The primary contributions of this research are summarized as follows:
\begin{enumerate}
    \item The CA-YOLO network is proposed, which introduces the MHSA module to enhance detail capture. It is specifically optimized for small target detection and innovatively designs the CFAM module to improve the efficiency of feature fusion, significantly improving the ability of small target detection.

    \item Inspired by biological positioning mechanisms, a new bionic pan-tilt tracking control strategy is designed. This strategy includes various mechanisms such as target center positioning and adaptive control parameter adjustment, effectively improving tracking accuracy and system robustness.

    \item The performance of the improved model is significant. The average accuracy on the COCO and VisDrone datasets is increased by 3.94\% and 4.90\% respectively, and the integration of the CA-YOLO algorithm and adaptive bionic pan-tilt technology has been experimentally verified to be effective.
\end{enumerate}



\section{related Work}
Target detection technology plays a pivotal role in applications such as autonomous driving and video surveillance, where real-time performance is critical. Convolutional Neural Network (CNN) based algorithms, particularly those employing deep learning, have demonstrated exceptional efficiency and accuracy in such scenarios. However, despite their advancements, the YOLO series algorithms face persistent challenges in detecting small targets and recognizing objects in complex backgrounds \cite{ref37}.

Several recent studies have attempted to address these limitations. C. Liu et al. proposed the YOLO-PowerLite model, based on YOLOv8n, which integrates BiFPN and employs the C2f\_AK module to optimize the detection of abnormal targets in power transmission lines. This approach reduces model parameters and computational requirements. Similarly, X. Liu et al. introduced the BGS-YOLO model, which combines GAM and SimC2f modules to significantly enhance road target detection performance, further showcasing the versatility and efficacy of BiFPN in diverse detection tasks \cite{ref3, ref5}.

Y. Liu et al. proposed YOLO-TS model with NWD loss, CIoU, StarBlock module, and Slimneck structure, aiming to boost small-target detection efficiency and accuracy \cite{ref4}. Likewise, Q. Xu et al. designed an intelligent road information collection and alarm system based on YOLOv3, which enhances detection accuracy and efficiency through improved feature extraction and attention mechanisms \cite{ref6}.

In construction and industrial applications, C. Liang et al. proposed the HRHD algorithm, which utilizes Transformer and RepConv to improve helmet detection accuracy and speed at construction sites. T. Zhang et al. introduced the GDM-YOLO model, enhancing the detection of steel defects while reducing computational complexity through the integration of SPDG, C2f-DRB, and MFEB modules \cite{ref7, ref8}.

Z. Sun et al. further reconstructed the YOLOv8 neck structure by incorporating BiFPN and SimSPPF modules to improve speed and accuracy, while the addition of the LSK-attention mechanism expanded the receptive field for enhanced detection precision \cite{ref9}. H. Wang et al., Q. Luo et al., and L. Wang et al. focused on improving the recognition of small targets in complex environments. These studies introduced various innovations, including loss functions, attention mechanisms, Deformable Convolutional Networks (DCN), small target detection heads, global attention mechanisms, and modules such as C2f-DSConv, DyHead dynamic detection heads, and the MPDIoU regression loss function \cite{ref10, ref11, ref12}.

While the YOLO series has achieved notable improvements through these enhancements—such as the integration of C2f\_AK, BiFPN, NWD, and related modules—it continues to face limitations in small target detection and handling complex backgrounds. Notably, these systems often fail to maintain tracking when targets move out of the field of view, highlighting a critical gap that remains unresolved.

In the field of biomimicry, biological vision systems have demonstrated excellent performance in target detection and stable tracking due to their high efficiency, robustness, and dynamic adaptability, providing key inspiration for the design of artificial vision systems. However, there are still many limitations in the current simulation studies of biological vision mechanisms, especially in the robustness and adaptability of the system in complex environments. Y. Deng et al., together with Q. Fu et al. and S. Wang et al. designed a vision platform system inspired by the structural and physiological mechanisms of the eagle eye. Y. Deng et al. designed and implemented a biological eagle eye vision platform for target detection, covering hardware configuration and image splicing to improve visual resolution and enhance target perception capabilities \cite{ref13}. Q. Fu et al. constructed a vision system incorporating a two-degree-of-freedom gimbal and cameras with varying focal lengths, which effectively improves the precision of target detection and optimizes tracking effects \cite{ref14}. S. Wang et al. designed a miniature vision system for small target detection, where cameras with different focal lengths are coordinated to achieve accurate detection of small targets \cite{ref15}.

H. Zou et al. constructed a three-dimensional kinetic eye control system model of the bionic eye based on the kinetic nerve pathway of human eyes. This model was applied to the control system of UAV onboard gimbal cameras, significantly enhancing the UAV's visual tracking performance \cite{ref16}.

L. Li et al. improved target detection performance by simulating the unique structural and functional characteristics of insect visual systems. Meanwhile, Wang et al. \cite{ref45}. developed a miniature vision system specialized for enhancing small target detection performance.

Although the above vision systems mentioned have improved the target detection accuracy, the existing system that simulates the biological vision mechanism is difficult to work stably and the target detection accuracy decreases dramatically when facing the actual complex environment with light changes, background interference, and variable-speed moving targets.

Pan-tilt tracking technology, an intelligent tracking approach based on computer vision, enables real-time identification and tracking of target objects' motion trajectories. However, structural limitations often result in unstable tracking speeds, challenges in addressing target occlusion, and inadequate scalability and anti-disturbance capabilities in complex environments \cite{ref38}.

Several advancements in pan-tilt tracking systems have been proposed. H. S. Lin et al. developed a YOLO-based target detection system combined with edge computing and an optical zoom pan-tilt camera, enhancing the remote monitoring capabilities of unmanned ground vehicles (UGVs) \cite{ref17}. Similarly, C. E. Lin et al. designed an unmanned aerial vehicle (UAV) system capable of accurately tracking ground targets through YOLO and pan-tilt control mechanisms \cite{ref18}.

G. C. Patru et al. introduced a cost-effective quadrotor UAV platform that integrates visual servoing with pan-tilt control to achieve precise target tracking and photography \cite{ref19}. W. Whitacre et al. proposed the SeaScan UAV system, equipped with a pan-tilt camera, which enables accurate tracking and photography of ground targets. This system accounts for factors such as UAV altitude, camera field of view, and geolocation tracking performance \cite{ref20}. 

Y. Lai et al. combined fuzzy control with Kalman filtering for efficient pan-tilt control \cite{ref21}, while L. Huang et al. utilized background update and fast model matching techniques with polar coordinate fuzzy control to achieve real-time target tracking \cite{ref22}. Additionally, A. Manecy et al. proposed a minimalist artificial eye featuring a lightweight pan-tilt system, optimized for precise positioning and tracking of small UAVs \cite{ref23}.

Despite the progress in real-time target tracking with pan-tilt systems, challenges persist, particularly in terms of stability and anti-disturbance performance. Moreover, most existing studies focus primarily on tracking algorithms, with limited exploration of control strategies specific to the pan-tilt mechanism itself.

To address the above problems, this paper presents a bionic stabilized positioning system based on CA-YOLO. The system not only integrates innovative improvements to the target detection algorithm, but also develops an innovative bionic gimbal control strategy, which is integrated with the bionic gimbal technology. This dual innovation aims to overcome the limitations of the traditional methods and enhance the robustness and accuracy of the system in dynamic complex environments.

\section{research method}
\subsection{Overall framework}
The design of the bionic pan-tilt system is inspired by the visual stability mechanisms observed in biological organisms, particularly the intricate structure of the human Vestibulo-Ocular Reflex \cite{ref39} (VOR), as illustrated in figure \ref{fig1}. The primary components of the bionic pan-tilt system are as follows: 

\begin{figure*}[!htbp]
\centering
\includegraphics[width=0.9\textwidth]{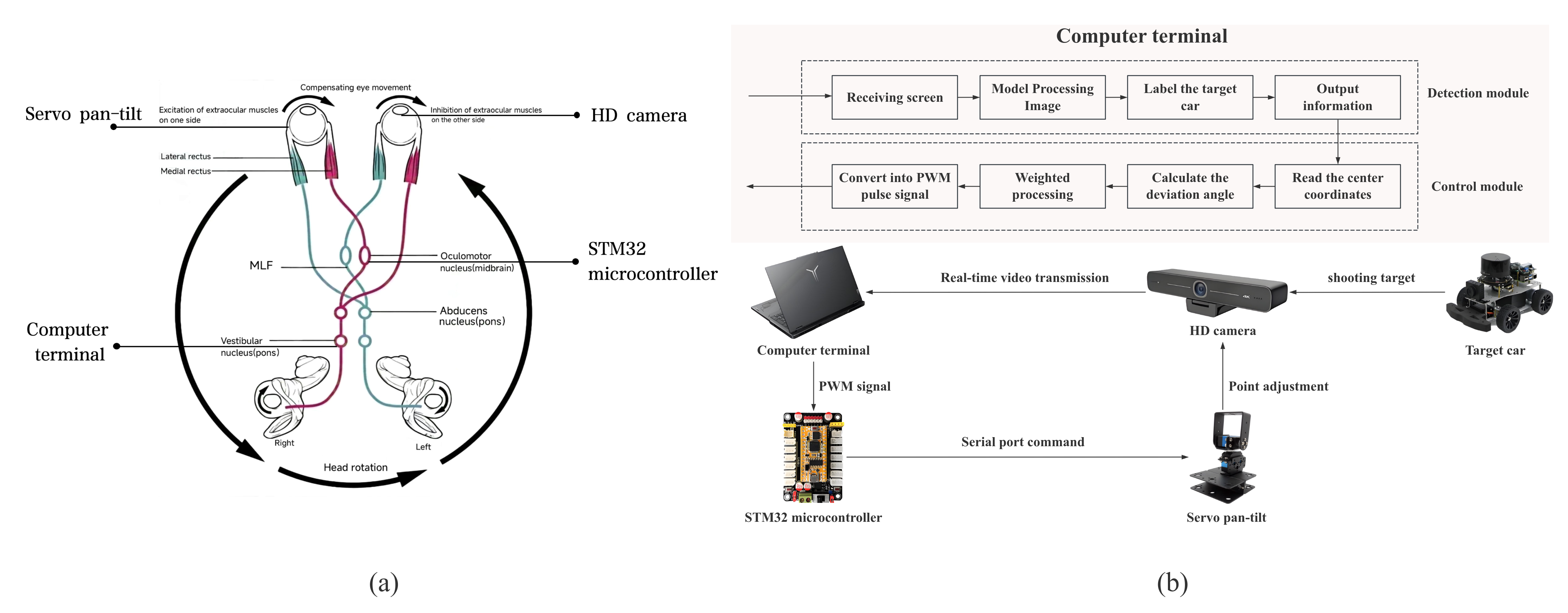}
\caption{(a) Showcased the construction of human Vestibular-Ocular Reflex (VOR), providing inspiration for the design of the bionic pan-tilt; (b) presents the overall framework of the bionic pan-tilt system, which includes a high - precision camera, a computer terminal, an STM32 single-chip microcomputer, and a servo pan-tilt. Multi-scale target detection and stable tracking are achieved through the CA-YOLO module.}
\label{fig1}
\end{figure*}

HD Camera: Simulating the function of the human retina, the camera serves as the "visual perception organ" of the system. It captures high-resolution images of the environment, providing essential raw data for subsequent image processing and analysis.

Computer Terminal: Acting as the system's "brain" and corresponding to the vestibular nucleus, the computer terminal processes the visual information captured by the camera. Connected to the camera via a USB interface, it performs image analysis and outputs control instructions based on the detected targets.

STM32 Microcontroller: Functioning as the system's "nerve center" and analogous to the oculomotor nucleus, the microcontroller receives processed information from the computer terminal. It transmits control signals via a serial communication interface to the servos of the pan-tilt, ensuring precise rotational control.

Servo Pan-Tilt: Emulating the role of extraocular muscles, the servo pan-tilt rotates the camera via microcontroller commands, with camera stability ensured by threaded-bolt mounting, mimicking human eye mechanics \cite{ref44}.

The overall framework of the bionic pan-tilt system is depicted in figure \ref{fig1}. Upon activation, the system immediately initializes the camera, which begins capturing real-time video streams of the target. These video streams are transmitted to the computer terminal's detection module, incorporating the CA-YOLO algorithm. By integrating the Multi-Head Self-Attention (MHSA) mechanism, a small target detection head, and the Characteristic Fusion Attention Mechanism (CFAM), the CA-YOLO module significantly enhances the performance of multi-scale target detection. The improved image processing algorithm identifies and labels targets within the video stream.

Once a target is identified, the control module calculates the angular deviation of the target from the screen center and converts this information into Pulse Width Modulation (PWM) signals via an optimized algorithm. These signals are transmitted to the servo pan-tilt to precisely adjust its rotation, ensuring the target remains centered in the field of view. Additionally, the PWM values are used to evaluate the system's target-tracking performance.

The system achieves efficient communication and coordination between modules through real-time video transmission and serial command exchanges. The seamless integration of the detection and control modules ensures high accuracy and real-time performance in target tracking. Furthermore, the system incorporates an automatic search function to recover tracking in cases of target loss, thereby maintaining continuity. This modular design offers a robust and efficient solution for target tracking in dynamic and complex environments.

\subsection{CA-YOLO Module: Improvement of Network Framework}
\subsubsection{CA-YOLO: Network Archivment} 
\begin{figure*}[htbp]
\centerline{\includegraphics[width=0.6\linewidth]{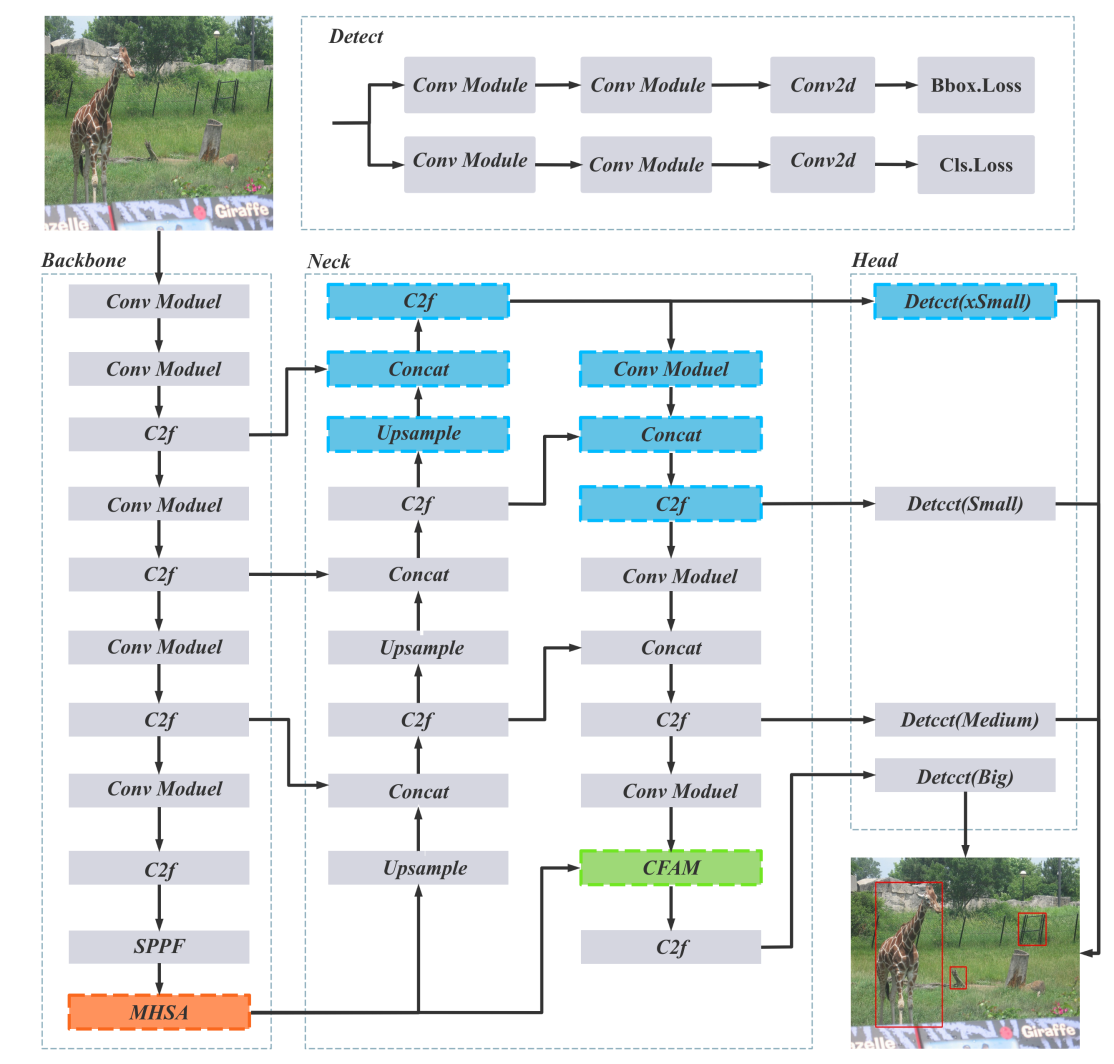}}
\caption{On the basis of YOLO, the CA-YOLO framework improves the performance of multi-scale object detection by placing MHSA after SPPF, adding a small object detection head (xSmall), and replacing some Concat modules with CFAM modules.}
\label{fig2}
\end{figure*}

Building upon YOLO network, the CA-YOLO framework introduces a series of innovative improvements, as shown in figure \ref{fig2}.
To improve the representation of feature maps and emphasize critical information, the Multi-Head Self-Attention (MHSA) mechanism is incorporated after the SPPF pooling layer. This modification significantly enhances the network's ability to detect and recognize targets at multiple scales.

To address the challenge of small target detection, a dedicated small target detection head (xSmall) is introduced into the network's original head structure. By leveraging the localized information from shallow feature maps, this addition improves detection accuracy for small targets and reduces missed detections. Furthermore, the original Concat module in the network's head structure is replaced with the CFAM module. This substitution facilitates fine-grained fusion of feature maps, enhancing the recognition performance for larger targets.

\subsubsection{Small Target Detection Head} 

The multi-channel visual mechanism of the biological eye gives it the ability to perceive different sizes and types of objects simultaneously. Some channels focus on capturing the contours and motions of large objects, while others are specialized in acquiring the details of small objects. This multi-channel synergistic effect significantly enhances the comprehensiveness and accuracy of visual perception \cite{ref46}.

\begin{figure*}[htbp]
\centerline{\includegraphics[width=0.8\linewidth]{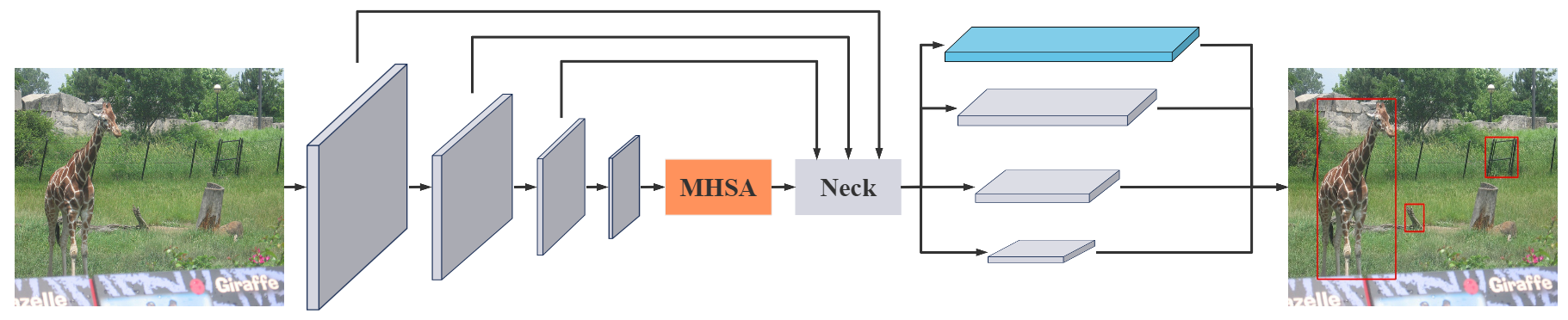}}
\caption{ Schematic diagram of the newly added small target detection head structure in CA-YOLO network. The higher-resolution detection layer retains more details of small targets, enhancing the accuracy of small target detection and reducing the leakage rate.}
\label{fig3}
\end{figure*}

To address the efficiency issues in small target detection caused by low pixel proportion and easy feature loss, this study draws on the strategies of global multi-level perception and dynamic region aggregation and innovatively introduces a dedicated small target detection head \cite{ref24,ref64}. By adding a detection layer with higher resolution to the shallow feature maps of the network, it can retain more detailed information, thereby improving the accuracy and robustness of detection. The added components of the framework are shown in  figure \ref{fig3}.

\subsubsection{CFAM Module} 

In complex natural environments, organisms can automatically adjust their attention allocation according to the target's importance or the level of potential threat. This attention allocation mechanism allows organisms to efficiently process critical information without ignoring potential threats in the surrounding environment \cite{ref47}. Inspired by this biological mechanism, to enhance the efficiency and effectiveness of feature fusion, this study introduces the Characteristic Fusion Attention Mechanism (CFAM) module as an innovative replacement for the traditional Concat module.

The CFAM module integrates an attention mechanism, which can perform refined regulation and fusion on feature maps of different scales, optimizing the multi-scale feature integration process. Compared with HR-FPN, which focuses on the scale problem of static small targets and adopts a feature fusion method based on hierarchical semantic characteristics \cite{ref63}, CFAM dynamically adjusts the fusion weight according to the target movement speed, taking into account the dynamic characteristics of the target. In contrast to the simple concatenation strategy of the Concat module, CFAM can accurately enhance key features and improve the model's adaptability to the variability of target sizes.

In the CFAM module, two feature maps,  \( X_0 \) and \( X_1 \), with identical channel dimensions but from different levels, are processed. A \( 1 \times 1 \) convolutional layer is applied to achieve nonlinear feature transformation. This operation enriches the representational capacity of the feature maps and enhances the model’s ability to capture complex inter-feature relationships, thereby increasing the efficiency and robustness of feature fusion and processing.

\begin{equation}
X_{0}^{\prime} = {adjust\_conv}(X_{0})
\end{equation}

The HFSM module enhances information interaction by concatenating multi-branch features and introducing an attention mechanism  \cite{ref40}. Inspired by this, the adjusted feature maps \(X_0'\) and \(X_1\) are concatenated along the channel dimension to form a new feature map \(X\_\text{concat}\). This new feature map is then processed by the Multi-Head Self-Attention (MHSA) mechanism.

\begin{figure}[t]
\centerline{\includegraphics[width=0.7\linewidth]{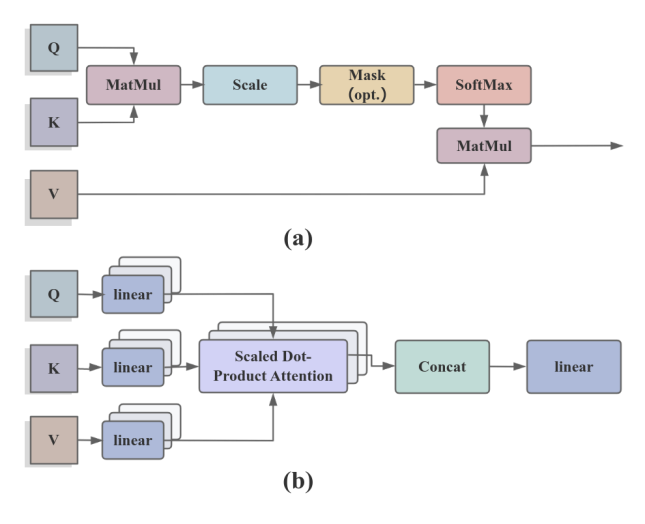}}
\caption{Decomposition diagram of the computational process of the multi-head self-attention mechanism (MHSA). (a) Illusttation of single-head attention computation: Q, K and V vectors are computed by matrix multiplication to compute similarity; (b) Presenting multi-head attention integration, Q, K and V are linearly transformed to generate representations for each head, and each head computes the single-head attention and then splices them together, and finally integrates them through the linear layer to get the MHSA output.
}
\label{fig4}
\end{figure}

MHSA captures long-range feature dependencies to improve small-target feature extraction efficiency \cite{ref25}.
The calculation process of single-head attention and the integration process of multi-head attention are shown in  figure \ref{fig4}.

Single-head attention calculation: The input query ($Q$), key ($K$), and value ($V$) vectors first perform an initial similarity calculation through matrix multiplication (MatMul).

\begin{equation}
{Attention}(Q, K, V) = {softmax}\left( \frac{QK^T}{\sqrt{d_k}} \right) V
\end{equation}

Where $Q$ represents the query matrix, $K$ is the key matrix, $V$ is the value matrix, and $d_{k}$ denotes the dimension of the key vector.

Multi-head attention integration: The input $Q$, $K$, and $V$ undergo independent linear transformations to generate separate representations for each attention head.

\begin{equation}
Q = W^Q X, \quad K = W^K X, \quad V = W^V X
\end{equation}

Where \(W^Q\), \(W^K\), \(W^V\) are learnable weight matrices, and \(X\) represents the input feature map.

Each head calculates single-head attention independently, extracting features from different subspaces. The outputs are concatenated to facilitate information sharing, followed by integration through a linear layer to produce the final output of MHSA.

\begin{equation}
{head}_{i} = {Attention}(QW_{i}^{Q}, KW_{i}^{K}, VW_{i}^{V})
\end{equation}
\begin{equation}
\text{MHSA}(Q, K, V) = {Concat}(head_1, \ldots, head_h) W^O
\end{equation}

Where \(\text{head}_i\) represents the attention output of the \(i\)-th head, and \(W^O\) another learnable weight matrix used to combine the outputs of the various heads. 

The processed \(X\_\text{concat}\) is then split into \(X_0'\_\text{weight}\) and \(X_1\_\text{weight}\), corresponding to the attention weights of \(X_0'\) and \(X_1\), respectively. Drawing on the idea of the DANTD model that dynamic discrimination is achieved by mining high-level features to adapt to scenario requirements \cite{ref66}., the weights are dynamically adjusted according to the target's movement speed. The specific mechanism is as follows:

The target movement speed is quantified by the normalized ratio of the displacement distance between the centers of the target detection boxes in two consecutive frames to the image size (a larger ratio indicates a higher speed). When the target moves fast, \(X_1\_\text{weight}\) (high-level feature weight, containing semantic and movement trend information) increases to strengthen global features; when the target moves slowly or is stationary, \(X_0'\_\text{weight}\) (low-level feature weight, containing small target details and texture information) increases to highlight details.

This dynamic adjustment mechanism allocates the proportion of weights in real time according to the target's movement state. The movement speed is positively correlated with the high-level feature weight and negatively correlated with the low-level feature weight, and the sum of the weights is always 1, which ensures the stability of feature fusion.

These weights are subsequently added to the original feature maps \(X_0\) and \(X_1\) to produce the weighted feature maps \(X_0'\_\text{sum}\) and \(X_1\_\text{sum}\), which further enhance the feature representation. This process allows the model to focus on the most relevant information within the image. 

\begin{equation}
X_{0}^{\prime} \_{sum}=X_{0}+X_{0}^{\prime} \_ {weight}
\end{equation}
\begin{equation}
X_{1} \_sum=X_{1} +X_{1} \_weight
\end{equation}

Finally, a deep cross-fusion is performed between the feature maps \(X_0'\_\text{sum}\) and \(X_1\_\text{sum}\) is performed.

\begin{equation}
X_{0}^{\prime} \_{result }=X_{1} \_sum+X_{0} 
\end{equation}
\begin{equation}
X_{1} \_result=X_{0}^{\prime} \_  {sum }+X_{1} 
\end{equation}

\begin{figure}[t]
\centerline{\includegraphics[width=0.95\linewidth]{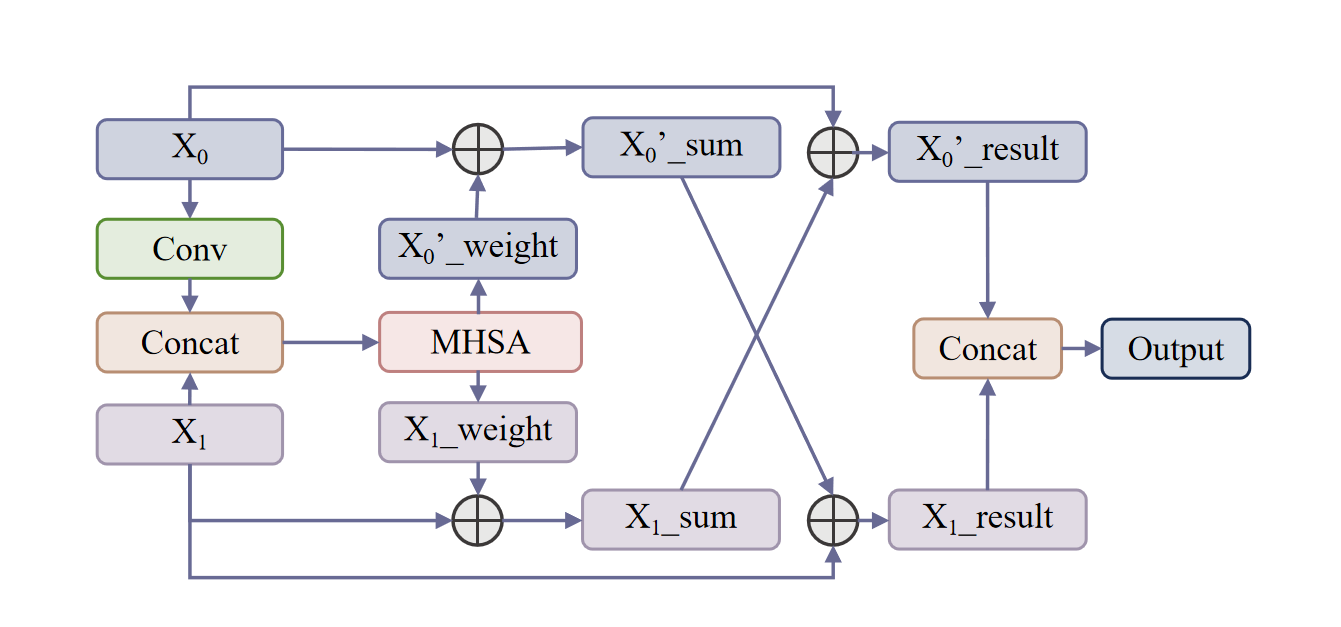}}
\caption{A hybrid feature fusion architecture that combines convolution and MHSA mechanism achieves information integration and output through weighted summation and feature concatenation.}
\label{fig5}
\end{figure}

The resulting fused feature maps, \(X_0' \_ \text{result}\) and \(X_1 \_ \text{result}\), are concatenated along the channel dimension to form the final output. This step significantly improves the information exchange between the feature maps, leading to a notable enhancement in the model’s image processing capabilities. The entire process is detailed in  figure \ref{fig5}.

In the experimental setup, the Concat module at the end of the network’s neck is replaced by the CFAM module. 
The modification preserves real-time efficiency while significantly enhancing large-target detection accuracy.

\subsection{Bio-Pan-Tilt Module: Precision Tracking System Inspired by Bionics}

This research introduces key optimizations to the control strategy of the bionic pan-tilt positioning system, including target center stabilization, decision boundary innovation, and adaptive tracking speed adjustment, all aimed at improving tracking efficiency. These strategies are further elaborated in the subsequent chapters.

\subsubsection{Bionic Pan-Tilt Tracking} 
Inspired by bionics, this study develops a bionic gimbal positioning system that captures real-time images via a high-precision camera, which are then transmitted to the terminal computer. The CA-YOLO module in the terminal analyzes these images and locates the target. Once the target is detected, the system calculates the deviation \((d_x, d_y)\) between the image center and the target center, and computes the horizontal ($H$) and vertical ($V$) deviation angles using the following formulas:
\begin{equation}
H=\left(\mathrm{d}_{\mathrm{x}} \times \theta_{\mathrm{x}}\right) / \mathrm{w}
\end{equation}
\begin{equation}
V=\left(\mathrm{d}_{\mathrm{y}} \times \theta_{\mathrm{y}}\right) / \mathrm{h}
\end{equation}

Where \(\theta_x\) and \(\theta_y\) represent the horizontal and vertical field of view angles of the camera, and \(w\) and \(h\) are the width and height of the image, respectively. 

The servo pan-tilt system is controlled by two servos, which rotate in the horizontal and vertical directions, mimicking the function of the extraocular muscles, and are driven by PWM signals. The initial PWM value is \(C\), corresponding to the neutral position of the servo. The PWM signal ranges from \(B\) to \(A\) (\(A > B\)), which covers a 270-degree rotation of the servo. The formula for converting the horizontal and vertical deviation angles into PWM is:
\begin{equation}
\text{PWM}_{H}=C-((A-B) \times H) / 270 
\end{equation}
\begin{equation}
\text{PWM}_{V}=C+((A-B) \times V) / 270
\end{equation}

Where \( \text{PWM}_H \) and \( \text{PWM}_V \) represent the PWM values for horizontal and vertical directions, respectively.

To ensure that the PWM value remains within the servo's physical working range, the system incorporates a calculation strategy that accounts for the saturation effect. This prevents mechanical damage and control failure, thereby maintaining the system’s stability and reliability. The following formula is used to adjust the PWM value:
\begin{equation}
\text{PWM}_{a}=\max \left(\min \left(\text{PWM}_{\text {c}}, \text{PWM}_{\max }\right), \text{PWM}_{\min }\right)
\end{equation}

Where \( \text{PWM}_{\text{min}} \) and \( \text{PWM}_{\text{max}} \) are the minimum and maximum PWM signal values, corresponding to the extreme rotation angles of the servo.

Additionally, the servo’s nonlinear dynamic characteristics, particularly during large-angle rotations, are considered in the control process \cite{ref33}. This is modeled using the following nonlinear dynamic equation:
\begin{equation}
\tau=J \frac{\mathrm{~d}^{2} \theta}{\mathrm{~d} t^{2}}+b \frac{d \theta}{d t}+K \theta+\tau_{\text {ext }}
\end{equation}

Where: 
\begin{itemize}
    \item \( \tau \) is the total torque acting on the system,
    \item \( J \) is the moment of inertia,
    \item \( \frac{d^2\theta}{dt^2} \)is the angular acceleration (rate of change of angular velocity),
    \item \( b \) is the damping coefficient,
    \item \( \frac{d\theta}{dt} \) is the angular velocity (current rotation speed),
    \item \( K \) is the stiffness coefficient,
    \item \( \theta \) is the current angular displacement,
    \item \( \tau_{\text{ext}} \)  is the external torque applied.
\end{itemize}

To ensure precise control, a PID controller is integrated within the servo, mimicking the feedback regulation mechanism found in organisms. The core of the PID controller consists of three components: proportional (\( P \)), integral (\( I \)), and derivative (\( D \)) actions, which collectively minimize the system error \cite{ref34}. The control formula is as follows: 
\begin{equation}
u(t)=K_{p} e(t)+K_{i} \int_{0}^{t} e(\tau) d \tau+K_{d} \frac{d e(t)}{d t}
\end{equation}

Where: 
\begin{itemize}
    \item \( K_p \) is the proportional gain, which determines the response strength to the current error,
    \item \( K_i \) is the integral gain, used to eliminate steady-state error by accumulating the error over time,
    \item \( K_d \) is the derivative gain, which predicts future errors and reduces overshoot and oscillation,
    \item \( e(t) \) is the error, or the difference between the setpoint and the actual output.
\end{itemize}

Through this closed-loop feedback system, precise control of the servo's rotation is achieved, allowing the camera angle to be adjusted and ensuring that the target remains centered in the image.

\subsubsection{Visual Center Positioning} 
In the tracking and positioning system, positioning the target at the center of the image is essential for enhancing both the stability and accuracy of tracking. This design principle is inspired by the human visual system, particularly the fovea in the retina, which is responsible for high-resolution processing of visual information.

To assess the importance of placing the target at the image center for improving tracking stability and accuracy, the Intersection over Union (IOU) of the model is compared and analyzed for targets located at the center and at the edges of the image. During the experiment, the target’s movement across different positions is precisely controlled, and a specific algorithm is employed to quantitatively evaluate the IOU of the tracking model. The formula for IOU is given by:

\begin{equation}
{IOU} = \frac{{Intersection}}{{Union}}
\end{equation}

Where the Intersection refers to the area of overlap between the two bounding boxes, and the Union represents the total area covered by both bounding boxes, including both the overlapping region and their respective unique areas.

\subsubsection{Visual Stability Optimization} 
The human visual system is capable of rapidly identifying and focusing on important targets within complex environments, dynamically adjusting attention based on the motion state of the target. In contrast, achieving precise center-point alignment of the servo in a pan-tilt system is a challenge due to various factors, including mechanical errors and environmental disturbances. To optimize this process, a decision boundary approach is proposed to simulate the human visual system's ability, thereby minimizing over-reactions to small angular deviations.

With the introduction of the decision boundary, when the angular deviation of the target is smaller than a predefined threshold, the system considers the target to be at the center of the image, reducing unnecessary adjustments. By implementing this decision boundary, the pan-tilt system can strike a balance between response speed and the stability of target tracking. This balance is crucial for enhancing tracking efficiency. To assess the impact of the decision boundary on target tracking efficiency, the following efficiency formula is used for evaluation:
\begin{equation}
\label{equ18}
\eta = \left(1 - \frac{T_w}{T_{w0}}\right) \times 100\%
\end{equation}

Where \(\eta\) represents the percentage improvement in tracking efficiency, \(T_w\) is the time taken to locate the target after introducing the decision boundary, and \(T_{w0}\) is the time required without the decision boundary.

\subsubsection{Adaptive Control Coefficient} 
When tracking variable speed or erratic targets, frequent control signals can induce vibrations in the gimbal due to inertia, which negatively impacts tracking stability and accuracy \cite{ref27}.

To address the issue of balancing stability and response speed in target tracking, the pan-tilt system adopts an adaptive control strategy. Drawing inspiration from the dynamic adjustment mechanism of human extraocular muscles, an intelligent coefficient K is introduced. The value of K is dynamically adjusted based on the motion state of the target, mimicking the natural response of the extraocular muscles to achieve stable target tracking. This strategy provides a more flexible and efficient solution for real-time target tracking through the dynamic optimization of adaptive control coefficients \cite{ref43,ref65}. This approach multiplies the angular information by \( K \), converts it into a PWM signal, and sends it to the microcontroller to control the rotation of the pan-tilt, similar to how the human visual system adapts to targets with varying speeds. The specific formula for adjusting \( K \)is as follows:
\begin{align}
\Delta d_{(t)} &= d_{(t)} - d_{(t-1)} \times \left(1 - K_{(t-1)}\right) \\
K_{(t)} &= K_{(t-1)} + \gamma \times \operatorname{sgn}\left(\Delta d_{(t)}\right) \\
\operatorname{sgn}\left(\Delta d_{(t)}\right) &= 
    \begin{cases} 
    1, & \Delta d_{(t)} \geq 0 \\
    -1, & \Delta d_{(t)} < 0 \\
    \end{cases} \\
K_{(t)} &= \min\left(\max\left(K_{(t)}, K_{\min}\right), K_{\max}\right)
\end{align}

Where \( K_{\min} \) and \( K_{\max} \) are the minimum and maximum values of the intelligent coefficient \( K \) respectively, \( d_{(t)} \) and \( K_{(t)} \) represent the distance between the target and the center of the image, and the value of \( K \) at time \( t \), respectively. \( d_{(t-1)} \), \( K_{(t-1)} \) denote the distance and the value of \( K \) at the previous time point \( (t-1) \). The coefficient \( \gamma \) , set to 0.1, controls the rate of change of \( K \), and \( \operatorname{sgn}(\Delta d_{(t)}) \) is the sign function.

The physical meaning of the above equations can be understood through the change in distance between the target in the image and the center of the frame: \(d_{(t-1)}\) is the distance between the target and the center of the frame in the previous frame; \(d_{(t-1)} \times (1 - K_{(t-1)})\) represents the remaining distance between the two centers after the frame center moves under the \(K_{(t-1)}\) value of the previous frame; \(d_{(t)}\) is the actual distance between the target and the frame center in the current frame. When \(\Delta d_{(t)} > 0\), it indicates that the actual distance is greater than the theoretical remaining distance, meaning that at the adjustment speed of the previous frame, the frame center cannot catch up with the target's movement in time, and thus the \(K_{(t-1)}\) value needs to be increased to speed up the response; when \(\Delta d_{(t)} < 0\), the \(K_{(t-1)}\) value should be decreased to avoid overshoot; when \(\Delta d_{(t)} = 0\), the actual distance completely matches the theoretical remaining distance, but to cope with the possible accelerated movement of the target in subsequent frames (predictive adjustment), the \(K_{(t-1)}\) value still needs to be moderately increased to avoid tracking lag in case of sudden speed changes. Meanwhile, the \(K_{(t)}\) value is confined to a reasonable range to prevent mechanical jitter or response lag. This mechanism achieves smooth tracking of variable-speed targets through dynamic optimization, echoing the adaptive adjustment mechanism of human extraocular muscles.

\begin{figure}[t]
\centerline{\includegraphics[width=0.7\linewidth]{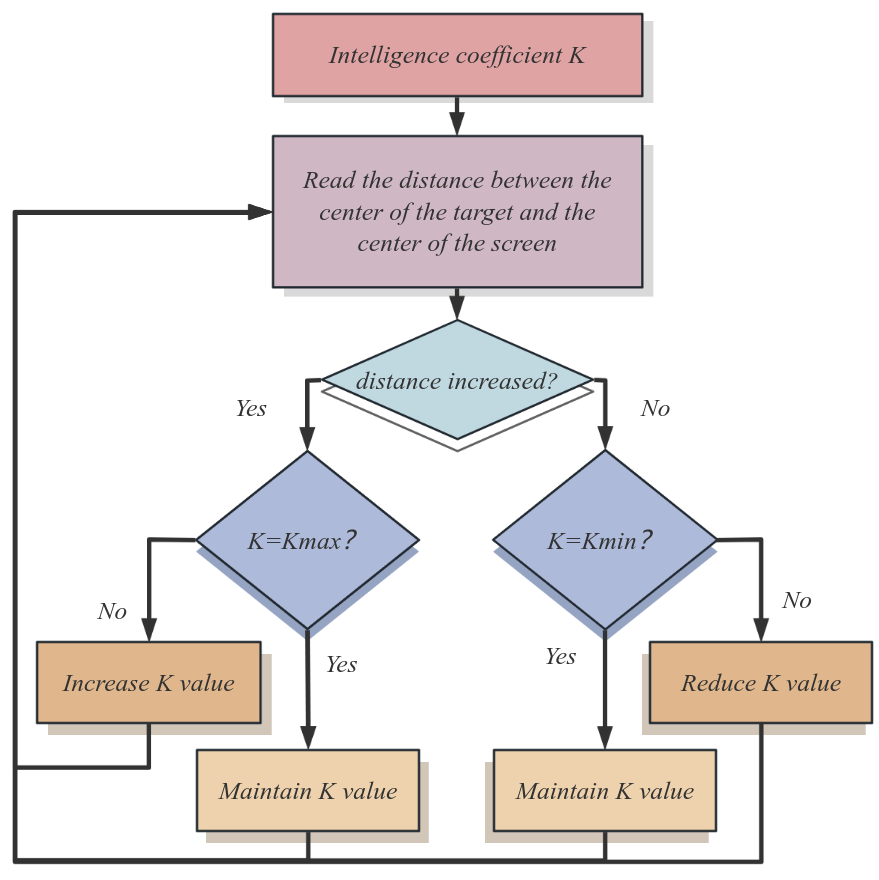}}
\caption{Flow chart of the dynamic adjustment mechanism for the intelligent coefficient K of the bionic pan-tilt. The K value is adjusted according to the change in distance between the target and the center of the frame in two adjacent frames.}
\label{fig6}
\end{figure}

The adjustment mechanism of the value of \( K \) is illustrated in  figure \ref{fig6}. Input the initial value of the intelligent coefficient \( K \), and the system detect whether the distance between the center point of the image and the target's center in two consecutive frames is increasing. If the distance increases,\( K \) is increased. Once \( K = K_{\max} \) and the distance continues to increase, \( K \) remains unchanged. Conversely, if the distance decreases, \( K \) is reduced. When \( K = K_{\min} \) and the distance continues to decrease, \( K \) remains unchanged. 
\subsubsection{Intelligent Target Recapture} 
In target tracking, the camera may temporarily lose the target due to occlusion, rapid movement, or other unforeseen circumstances. To address this issue, an automatic search function is introduced, simulating the adaptability and recovery capabilities of organisms in complex environments. This function is activated when the target is lost, causing the pan-tilt to rotate back and forth between the two extreme horizontal positions until the target is detected again.

Additionally, this research proposes a chronic search strategy based on historical data analysis, which prioritizes searching in areas with high probability of target presence. This strategy mirrors the way organisms utilize memory and experience to enhance search efficiency in familiar environments. The chronic search is focused near the pan-tilt’s initial position, under the assumption that this area corresponds to regions where the target has historically been active.

By dynamically adjusting the search speed, the pan-tilt allocates more time to areas where the target is most likely to be found, optimizing search efficiency and increasing the likelihood of target recovery. This probabilistic, history-based approach enhances the pan-tilt's adaptability to environmental changes, providing an effective solution for target tracking in complex and dynamic environments. 
\section{Experimental results and analysis}
\subsection{CA-YOLO Module} 
\subsubsection{Experimental Environment} 
The experimental setup was conducted on a Windows operating system. The hardware configuration included an Intel(R) Core(TM) i9-14900HX CPU and an NVIDIA GeForce RTX 4090 GPU. The deep learning framework used was PyTorch, with Python version 3.10.15 and CUDA version 12.6. YOLOv8n served as the baseline model. The hyperparameters were set as follows: a batch size of 32, 8 workers, 500 epochs, and an initial learning rate of 0.01.
\subsubsection{Data Sets} 

In the training session of this research, four datasets were utilized, including two publicly available datasets: the COCO dataset and the VisDrone dataset; as well as two customized datasets: the AGV dataset and the UAV dataset.

The COCO dataset serves as a pivotal resource in computer vision, offering a large-scale dataset for object detection, image segmentation, and image caption generation tasks \cite{ref61}. The VisDrone dataset, a prominent large-scale dataset from UAV viewpoints, is widely applied to target detection and tracking research in domains such as intelligent transportation, disaster response, urban planning, and agricultural monitoring \cite{ref62}.

The AGV dataset comprises 1,258 images of carts with varying postures, where each cart has dimensions of 271 mm × 189 mm × 151 mm. The UAV dataset contains 1,029 images of unmanned aerial vehicles in different attitudes, with UAV dimensions measuring 171 mm × 245 mm × 62 mm. Both customized datasets focus on a single target class. All images are randomly partitioned into training and validation sets at a 7:3 ratio.

In this study, all images were acquired at a resolution of 640 × 640. Small targets were defined as those occupying less than 0.12\% of the total image pixels, following the definition established by SPIE \cite{ref60}. Building on this, the study classifies targets occupying less than 0.1\% of the image area as small targets to adapt to the characteristics of high-resolution imagery. This criterion effectively differentiates small targets from large ones, providing a precise foundation for subsequent research. Based on this classification, the validation set is further subdivided into a small-target validation set and a non-small-target validation set to support accurate subsequent analysis.

\subsubsection{Evaluation Metrics} 
To evaluate the performance of the proposed model, the following metrics were used: Precision (P), Recall (R), Mean Average Precision (mAP), Frames per Second (FPS), Giga Floating-Point Operations per Second (GFLOPs), and the size of the weight file (MB).
\subsubsection{Model Testing and Evaluation} 
Two comparative experiments were conducted on the COCO dataset to demonstrate the effectiveness and superiority of the CA-YOLO model: 
\begin{enumerate}
    \item Comparison with various versions of YOLO.
    \item Comparison with other state-of-the-art detection models.
\end{enumerate}

To ensure consistency and fairness, all test images were resized to an input resolution of 640 × 640 pixels, with an IOU threshold of 0.7 and a confidence threshold of 0.001. The experimental results are summarized in Table~\ref{tab1}, presenting comparisons with different versions of YOLO and other advanced detection models.

\begin{table*}[!t]
\caption{Performance comparison of the CA-YOLO Model with different versions of YOLO and other advanced detection models on COCO dataset.}
\label{tab1}
\centering
\begin{tabular}{ccccccc}
\toprule
Model & F1 & mAP50 & mAP50-95 & GFLOPs & Params M & Model Size MB \\
\midrule
YOLOv8n & 0.52 & 51.46 & 36.33 & 8.90 & 3.16 & 6.27 \\
YOLOX-tiny \cite{ref48}& 0.53 & 52.80 & 33.80 & 7.63 & 5.06 &	19.40 \\
YOLOv9t \cite{ref49}& 0.53 & 52.20 & 37.60 & 8.20 & \textbf{1.99} & \textbf{4.73} \\
YOLOv3 (Darknet-53) \cite{ref50}& 0.55 & 49.70 & 28.80 & 57.15 &	61.95 & 236.00 \\
YOLO11n \cite{ref51}& 0.55 & 54.60 & 39.00 & \textbf{6.50} &2.50 &5.35 \\
CenterNet (ResNet-18) \cite{ref52}& 0.50 & 46.10 & 29.50 & 14.35 & 14.44 & 55.10 \\
DETR (ResNet-50) \cite{ref53}& 0.48 & 54.00	& 34.30 & 28.14 & 41.58 & 564.00 \\
Faster R R-CNN (ResNet-50) \cite{ref54}& 0.55 & 48.70 & 32.10 & 68.07 & 41.75 & 159.00 \\
FSAF (ResNet-50) \cite{ref55}& 0.53 & 47.80 & 32.00 & 60.26 &	36.42 & 139.00 \\
Mask R R-CNN (ResNet-50) \cite{ref56}& 0.55 & 46.70 & 29.50 & 127.46 & 44.40 & 169.00 \\
RetinaNet (ResNet-50) \cite{ref57}& 0.52 & 44.60 & 30.10 & 69.92 & 37.97 & 145.00 \\
SSD \cite{ref58}& 0.53 & 46.10 & 28.90 & 98.63 & 36.04 & 137.00 \\
CA-YOLO & \textbf{0.56} & \textbf{55.40} & \textbf{39.52} & 18.60 & 4.46 & 8.88 \\
\bottomrule
\end{tabular}
\end{table*}

To more intuitively demonstrate the improvement in small-target detection capability, this study conducted comparative experiments between the original model and the improved model on the VisDrone dataset—specifically designed for small-target detection. The experimental results are summarized in Table~\ref{tab2}.

\begin{table*}[!t]
\caption{Performance Comparison between the CA-YOLO Model and YOLOv8n Model on the VisDrone Dataset}
\label{tab2}
\centering
\begin{tabular}{ccccccc}
\toprule
Model & F1 & mAP50 & mAP50-95 & GFLOPs & Params M & Model Size MB \\
\midrule
YOLOv8n & 0.38 & 33.50 & 19.44 & 8.20 & 3.01 & 5.98 \\
CA-YOLO & 0.42 & 38.40	& 23.06 & 13.50 & 4.04 &8.07 \\

\bottomrule
\end{tabular}
\end{table*}

The ablation experiments for the model proposed in this paper were conducted on the COCO dataset under identical environmental conditions. The results are shown in Table~\ref{tab3}.

\begin{table*}[!t]
\caption{Comparison of the impacts of different improvement methods (MHSA, small target detection head, CFAM module) on the model performance in CA-YOLO models}
\label{tab3}
\centering
\begin{tabular}{cccccccccc}
\toprule
Baseline & MHSA & Small detect & CFAM & F1 & mAP50 & mAP50-95 & FPS & GFLOPs & Model Size MB \\
\midrule
YOLOv8n &  &  &  & 0.52 & 51.46 & 36.33 & \textbf{130} & \textbf{8.90} & \textbf{6.27} \\
YOLOv8n & $\checkmark$ &  &  & 0.54 & 53.39 & 37.76 & 125 & 9.00 & 6.65 \\
YOLOv8n &  & $\checkmark$ &  & 0.54 & 53.03 & 37.72 & 120 & 17.70 & 6.89 \\
YOLOv8n & $\checkmark$ &  & $\checkmark$ & 0.54 & 52.80 & 37.49 & 105 & 9.50 & 7.90 \\
YOLOv8n & $\checkmark$ & $\checkmark$ &  & 0.55 & 54.89 & 38.91 & 105 & 17.90 & 7.25 \\
YOLOv8n & $\checkmark$ & $\checkmark$ & $\checkmark$ & \textbf{0.56} & \textbf{55.40} & \textbf{39.52} & 100 & 18.60 & 8.88 \\
\bottomrule
\end{tabular}
\end{table*}

This study aims to enhance detection accuracy during target tracking and address the limitation of single-target scenes. To this end, the original YOLOv8 model and CA-YOLO model weights trained in Table~\ref{tab3} serve as pre-training weights, with training experiments conducted on the AGV dataset and UAV dataset, respectively. All experiments are performed under unified environmental conditions to ensure fairness and consistency of results. The relevant findings are presented in Tables~\ref{tab4} and ~\ref{tab5} where All Val denotes the overall validation set, Small Val represents the small-target validation set, and No-Small Val indicates the non-small-target validation set.

\begin{table*}[!t]
\caption{Comparison between the performance of YOLOv8n and CA-YOLO models on AGV dataset}
\label{tab4}
\centering
\begin{tabular}{cccccc}
\toprule
Model & mAP50 & mAP50-95 & F1 & Model Size MB \\
\midrule
YOLOv8n(All Val)& 95.10 & 73.10 & 0.92 & 6.29 \\
YOLOv8n(Small Val)& 93.30 & 66.50 & 0.89 & 6.29 \\
YOLOv8n(No-Small Val)& 99.20 & 83.40 & 0.99 & 6.27 \\
CA-YOLO(All Val)& 99.30 & 77.00 & 0.99 & 8.90 \\
CA-YOLO(Small Val)& 99.50 & 74.10 & 0.99 & 8.91 \\
CA-YOLO(No-Small Val)& 99.50 & 84.90 & 0.99 & 8.89 \\
\bottomrule
\end{tabular}
\end{table*}

\begin{table*}[!t]
\caption{Comparison between the performance of YOLOv8n and CA-YOLO models on UAV dataset}
\label{tab5}
\centering
\begin{tabular}{cccccc}
\toprule
Model & mAP50 & mAP50-95 & F1 & Model Size MB \\
\midrule
YOLOv8n(All Val)& 96.00 & 71.20 & 0.93 & 6.31 \\
YOLOv8n(Small Val)& 93.20 & 57.10 & 0.89 & 6.30 \\
YOLOv8n(No-Small Val)& 99.50 & 88.00 & 0.99 & 6.30 \\
CA-YOLO(All Val)& 97.60 & 73.10 & 0.96 & 9.00 \\
CA-YOLO(Small Val)& 95.70 & 60.50 & 0.93 & 8.95 \\
CA-YOLO(No-Small Val)& 99.50 & 88.70 & 0.99 &  8.99 \\
\bottomrule
\end{tabular}
\end{table*}

\begin{figure*}[t]
\centerline{\includegraphics[width=0.95\linewidth]{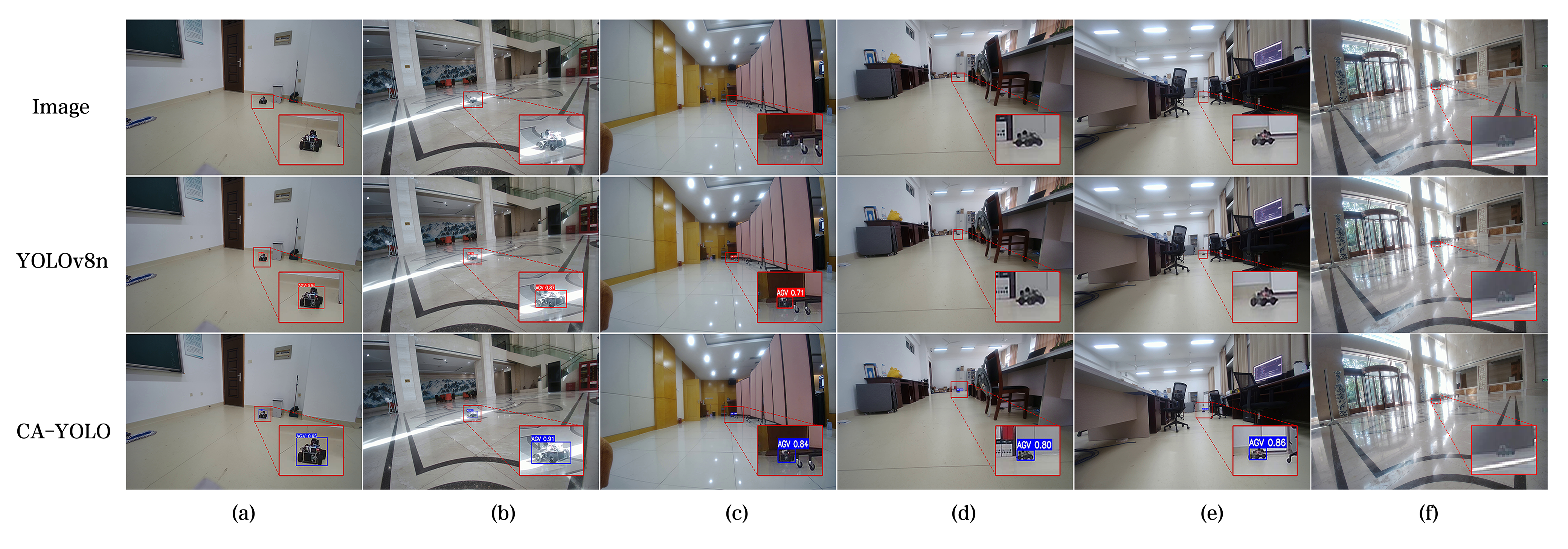}}
\caption{Comparison of AGV detection results between YOLOv8n and CA-YOLO models in different scenarios, presenting the two models’ detection outcomes on various types of AGV-containing images (with AGV-to-camera distances ranging from 2–10 meters). The results highlight the detection confidence and robustness advantages over the CA-YOLO model compared to the YOLOv8n model when dealing with long-distance or small-sized targets. }
\label{fig7}
\end{figure*}

\begin{figure*}[t]
\centerline{\includegraphics[width=0.95\linewidth]{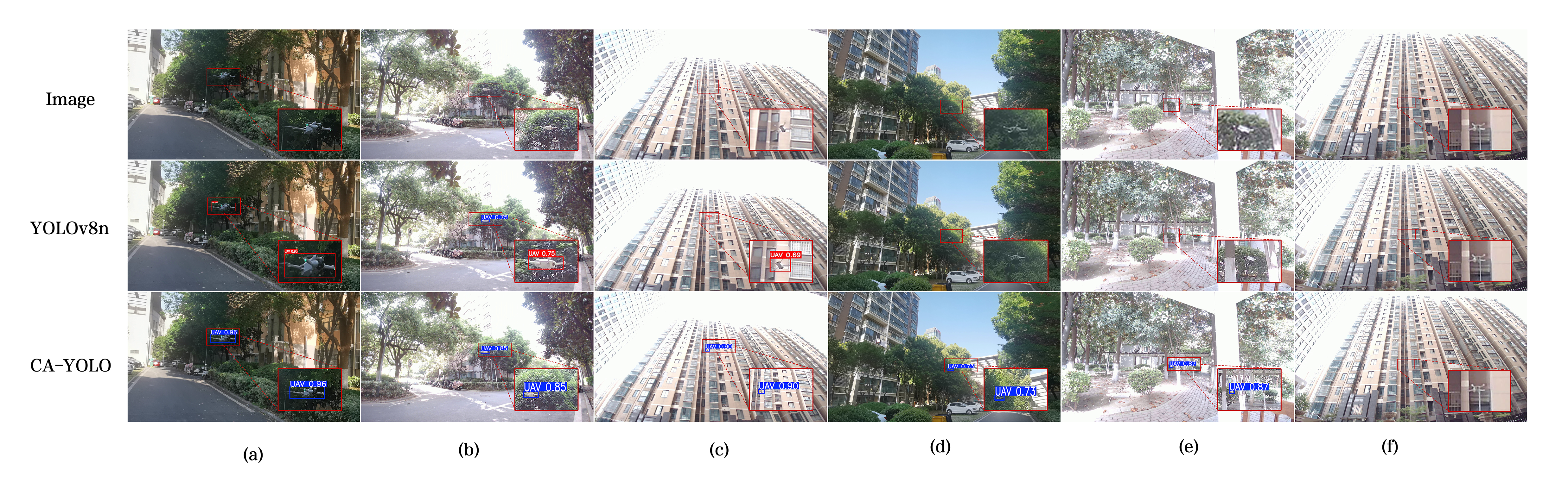}}
\caption{Comparison of UAV detection results between YOLOv8n and CA-YOLO models across different scenarios. The detection results of the two models on various types of images containing UAVs  (with UAV-to-camera distances ranging from 2–30 meters) are presented, highlighting the advantages of the CA-YOLO model over the YOLOv8n model in terms of detection confidence and robustness when dealing with long-distance or small-size targets.}
\label{fig14}
\end{figure*}

The two models are validated on various types of target-containing images, with some results shown in figure \ref{fig7}and \ref{fig14}. These results demonstrate the effectiveness of YOLOv8n and CA-YOLO models in detecting the desired targets in different scenarios.

\subsubsection{Conclusions and Analysis} 

The CA-YOLO model proposed in this study demonstrates remarkable performance in target detection tasks, particularly in enhancing the robustness of small target detection. This is achieved through several key improvements: the integration of a Multi-Head Self-Attention mechanism (MHSA), optimization of the small target detection head, and the incorporation of the CFAM module. The MHSA captures long-range dependencies to significantly improve the extraction of small target features. Meanwhile, the small target detection head leverages shallow feature map information for precise localization, effectively boosting the accuracy and robustness of detecting small targets. The CFAM module introduces an attention mechanism to achieve fine-grained fusion of feature maps. It selectively emphasizes key features, enhances the model’s adaptability to varying target sizes, and further elevates detection performance for large targets.

The results in Table~\ref{tab1} demonstrate that CA-YOLO outperforms other models on the COCO dataset, exhibiting its superiority in accuracy, efficiency and model size, and demonstrating excellent comprehensive performance. The results in Table~\ref{tab2} provide more evidence of the improved small target detection capability, and the results in Table~\ref{tab3} validate the contribution of the individual improvement methods in the model to the performance improvement, while maintaining a low computational consumption, proving the effectiveness of the improvement methods. Tables~\ref{tab4} and Tables~\ref{tab5} show that CA-YOLO performs significantly better than YOLOv8n on the AGV and UAV dataset, achieving higher detection accuracy on both the overall validation set, the small target validation set and the non-small target validation set, while consuming a reasonable amount of resources, proving its superiority.

Although the GFLOPs of CA-YOLO are higher than those of YOLOv8n, the model controls costs through lightweight strategies such as local window attention, limiting the position of the small target detection head, and 1×1 convolution for channel compression. In our experimental environment, its FPS still reaches 100, and the model size is 8.88 MB, which meets the requirements of real-time detection and edge deployment. The actual benefit of this trade-off between accuracy and cost far outweighs the moderate increase in computing resources.

The experimental results in figure \ref{fig7} and figure \ref{fig14} show that the CA-YOLO model exhibits stronger robustness than YOLOv8n in target detection tasks: both can recognize targets at close range, but as the distance increases, the confidence of YOLOv8n decreases significantly, and it is prone to missed detections. In contrast, CA-YOLO can not only recognize YOLOv8n-missed targets, but also has greater confidence when both detect targets, with prominent advantages especially in long-distance or small-sized targets. However, both models fail to detect when the target is under strong light or backlight interference, indicating that there is a common limitation in scenarios where extreme low pixels are combined with complex environments. This not only verifies the effectiveness of CA-YOLO in conventional scenarios but also provides a direction for subsequent optimization in extreme scenarios.

\subsection{Bio-Pan-Tilt Module} 
\subsubsection{Visual Center Positioning} 
In the experimental setup, the camera was fixed at a predetermined position to maintain a constant field of view. The target followed a straight-line trajectory from the left side to the right side of the camera's field of view, maintaining a constant distance (either 4 meters or 6 meters) from the camera. The central area of the frame was defined as the region around the center of the image, with both the height and width accounting for one-third of the frame’s resolution. The remaining area was classified as the edge area. Fifteen images were randomly captured during the target’s movement within both the central and edge areas for subsequent IOU analysis.

\begin{figure}[t]
\centerline{\includegraphics[width=1.0\linewidth]{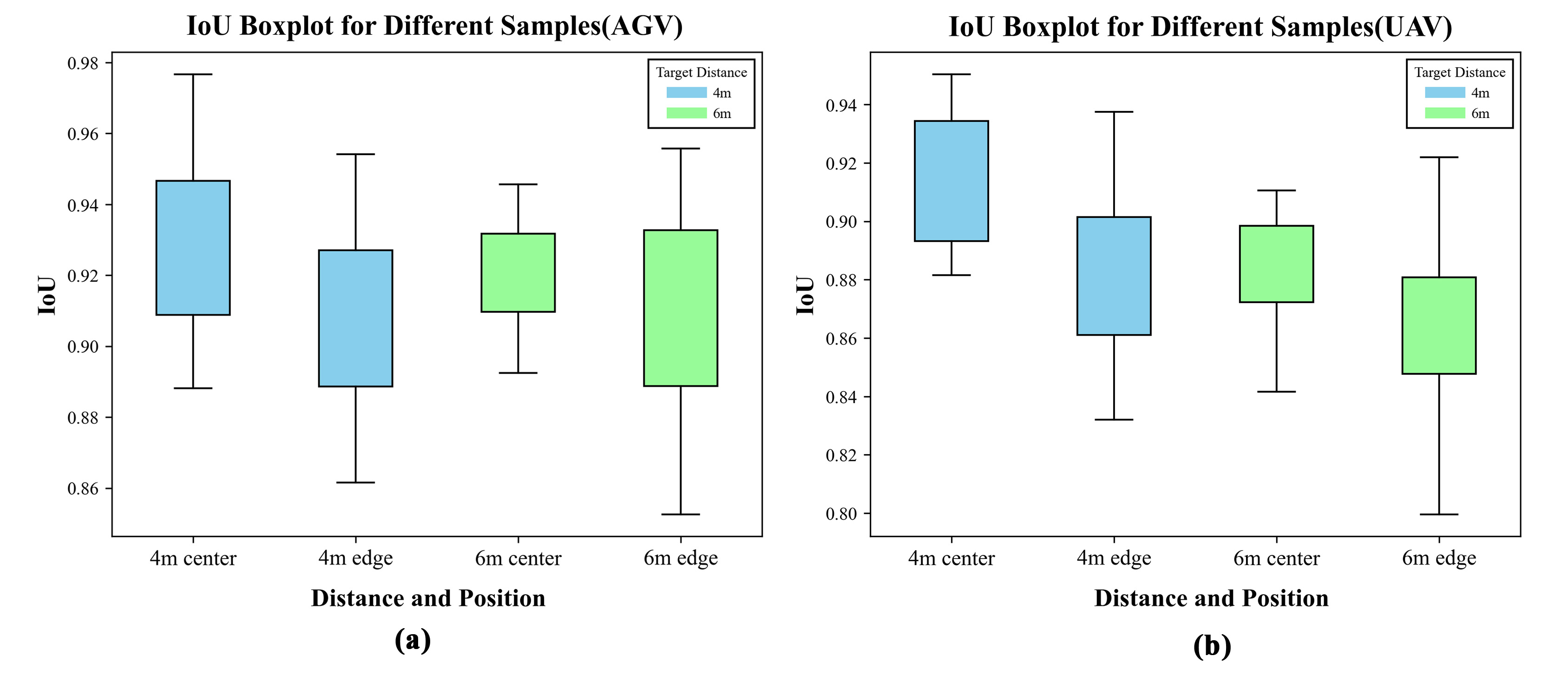}}
\caption{(a) and (b) present the box-and-line plots of the IoU performance of the target model at different distances and locations for the AGV and UAV, respectively. Target images are randomly captured at the center and edge of the frame at 4 m and 6 m. The IoU is analyzed to compare the performance of the four samples (4m-center, 4m-edge, 6m-center, 6m-edge) to detect the importance of target centering for optimal tracking}
\label{fig15}
\end{figure}

The experimental results are shown in figure \ref{fig15} and Tables~\ref{tab6}. Notably, the intersection over Union (IoU) is significantly higher when the target positioned at the center of the frame compared to when it is located at the edges, confirming the importance of keeping the target centered to optimize the tracking performance. This phenomenon was consistently verified for both 4-meter and 6-meter distance settings. In addition, the experimental results indirectly emphasize the critical role of the bionic gimbal in maintaining target localization, as it can dynamically adjust the camera view angle to keep the target centered, thus improving tracking stability and accuracy.

\begin{table*}[!t]
\caption{Performance comparison between the performance of YOLOv8n and CA-YOLO models on UAV dataset}
\label{tab6}
\centering
\begin{tabular}{cccccc}
\toprule
Target & 4m center Average IOU & 4m edge Average IOU & 6m cneter Average IOU & 6m edgeAverage IOU \\
\midrule
AGV & 0.93 & 0.91 & 0.92 & 0.90 \\
UAV & 0.91 & 0.88 & 0.88 & 0.86 \\
\bottomrule
\end{tabular}
\end{table*}

\subsubsection{Visual Stability Optimization} 
A decision boundary was set so that when the angular deviation between the target and the center of the image was less than 2 degrees, the system considered the target to be at the center and refrained from making excessive adjustments.

In the experiment, the target was placed at specific points within the camera’s field of view to observe the tracking behavior. The time required to locate the target with and without the decision boundary was compared to evaluate the impact of the decision boundary on tracking efficiency.
\begin{figure}[t]
\centerline{\includegraphics[width=0.95\linewidth]{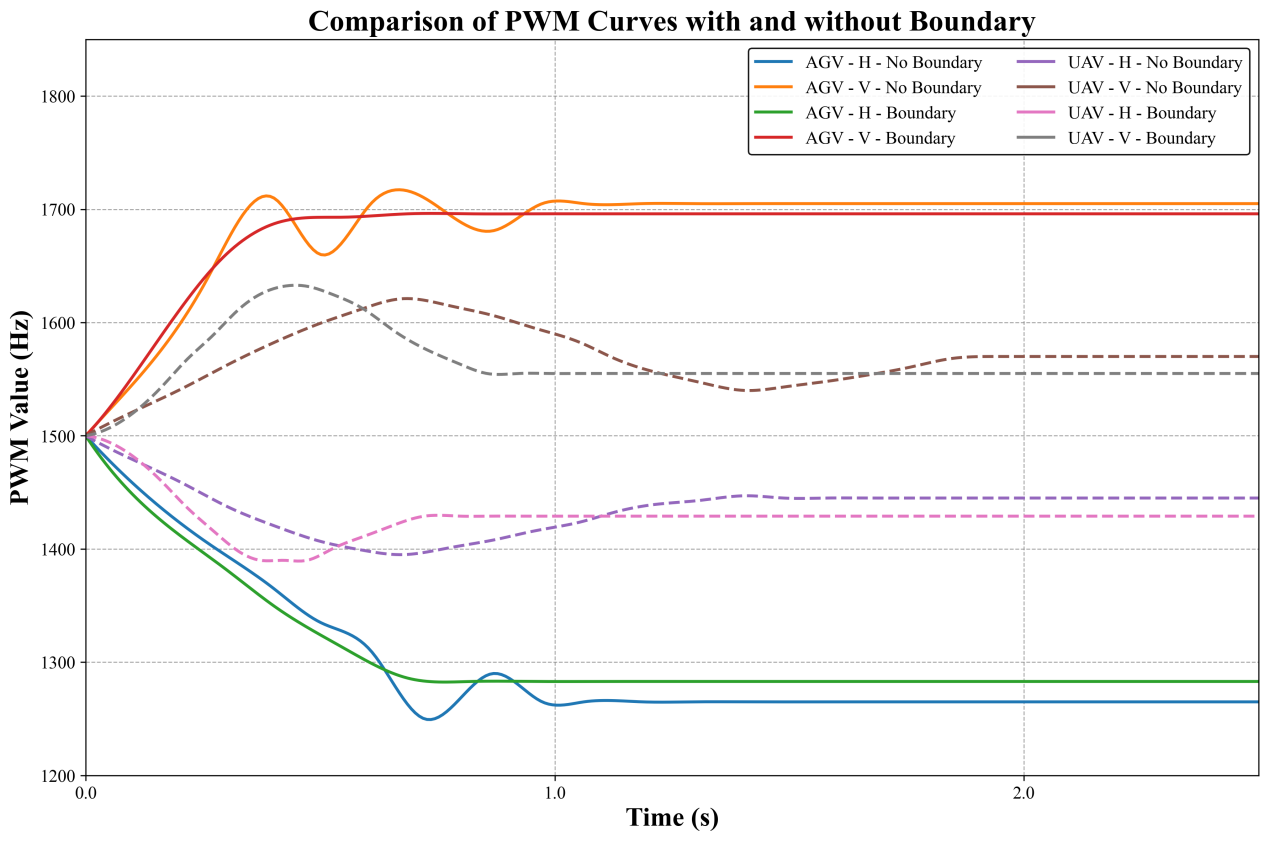}}
\caption{It shows the impact of decision boundaries on the tracking efficiency of different targets (where H stands for horizontal direction and V for vertical direction). Among them, the solid lines correspond to AGV, and the dashed lines correspond to UAV}
\label{fig16}
\end{figure}

The experimental results are shown in figure \ref{fig16}. Without the decision boundary, the gimbal continuously corrects for minor changes in the target angle, causing slight jerking and reduces gimbal stability. After the introducing the decision boundary, target positioning speed is significantly accelerated: the time for the AGV and the UAV to locate to the center is reduced from 0.973 seconds and 1.842 seconds to approximately 0.820 seconds and 0.855 seconds, respectively, as calculated by Equation (\ref{equ18}), with efficiency improvements of \(\eta = 15.725\%\) and \(\eta = 53.593\%\) respectively. The role of setting decision boundaries in improving the efficiency of target tracking is experimentally verified.

\subsubsection{Adaptive Control Coefficient} 
The intelligent coefficient \(K\) was introduced and dynamically adjusted based on the changes in the distance between the target and the center of the image in each frame. As the target moved farther away, increasing\(K\) accelerated the response; as the target approached, reducing \(K\) slowed the response \cite{ref27}.The value of \(K\)was constrained within the range of 0.2 to 0.6, i.e., \( K_{\min} = 0.2 \) and \( K_{\max} = 0.6 \), to ensure that the response speed was neither too slow, which could lead to target loss, nor too fast, which could cause rigid movement of the pan-tilt.

In this experiment, the decision boundary was set to 2 degrees, and the target was positioned within the camera’s field of view. Once the target was stable in the central area, it was controlled to start accelerating until it reached the predetermined endpoint. Two strategies were adopted for target tracking: setting a fixed parameter (with \(K=0.6\) ) and introducing the adaptive variable-speed control via \(K\) . These strategies simulated dynamic changes in actual tracking scenarios, allowing the evaluation of the response speed and tracking accuracy of the pan-tilt.
\begin{figure}[t]
\centerline{\includegraphics[width=0.95\linewidth]{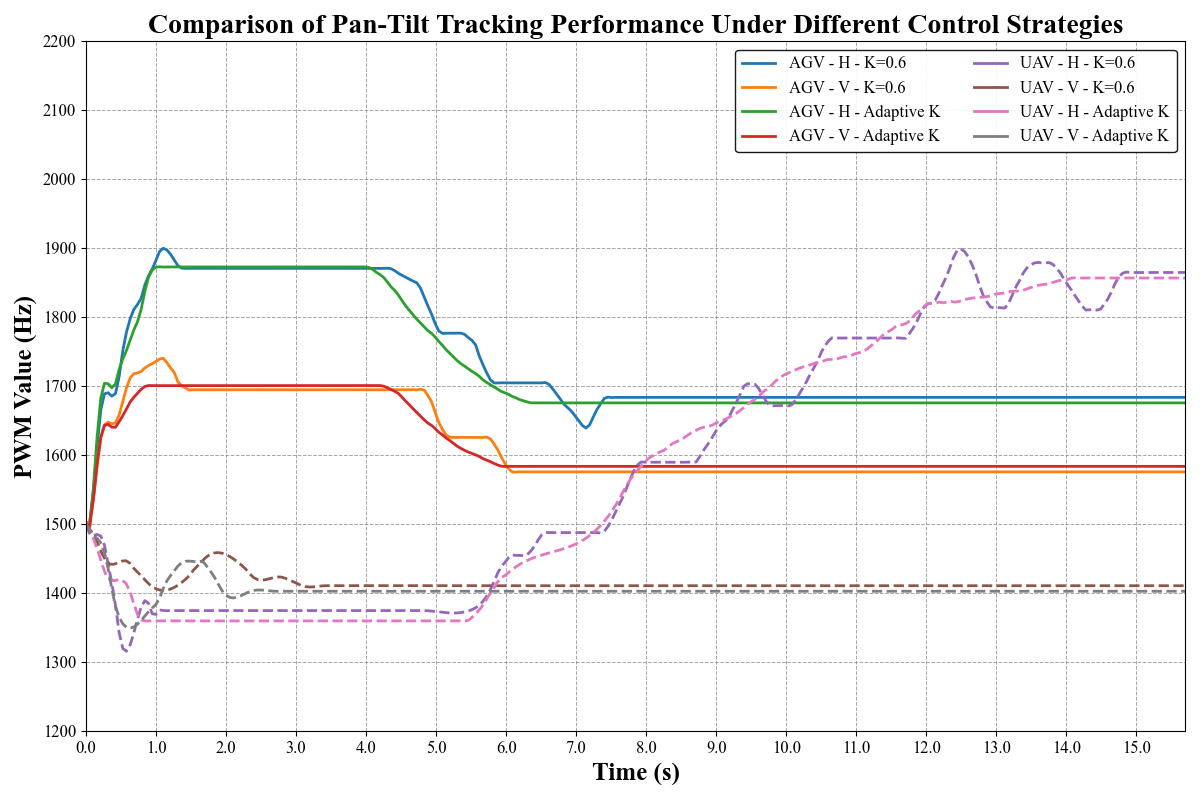}}
\caption{It shows the pan-tilt tracking performance between the fixed control coefficient (K=0.6) and the adaptive control strategy (where H represents the horizontal direction and V represents the vertical direction). Among them, the solid lines correspond to AGV, and the dashed lines correspond to UAV}
\label{fig17}
\end{figure}

The experimental results are presented in  figure \ref{fig17}. By comparing the changes in PWM values under both fixed and adaptive variable-speed control strategies, the tracking performance of the pan-tilt was analyzed. The results indicate that, with fixed parameters, the pan-tilt’s response to the target’s variable speed was not smooth, which could lead to a decrease in tracking accuracy. In contrast, the adaptive variable-speed control strategy significantly improved the smoothness of the pan-tilt’s movement by adjusting \(K\) in real-time, thus optimizing target tracking performance.

\subsubsection{Conclusions and Analysis} 
After conducting a series of experiments, the importance of keeping the target at the center of the image for enhancing tracking performance was confirmed, and this advantage was observed at both 4-meter and 6-meter distances. The introduction of the decision boundary notably improved the tracking efficiency of the pan-tilt, minimizing unnecessary adjustments. The adaptive variable-speed control strategy made the pan-tilt's response smoother and more natural, effectively handling the challenges of variable-speed target movement. The integration of these strategies substantially improved the stability and accuracy of the pan-tilt tracking system.

\begin{figure}[t]
\centerline{\includegraphics[width=0.8\linewidth]{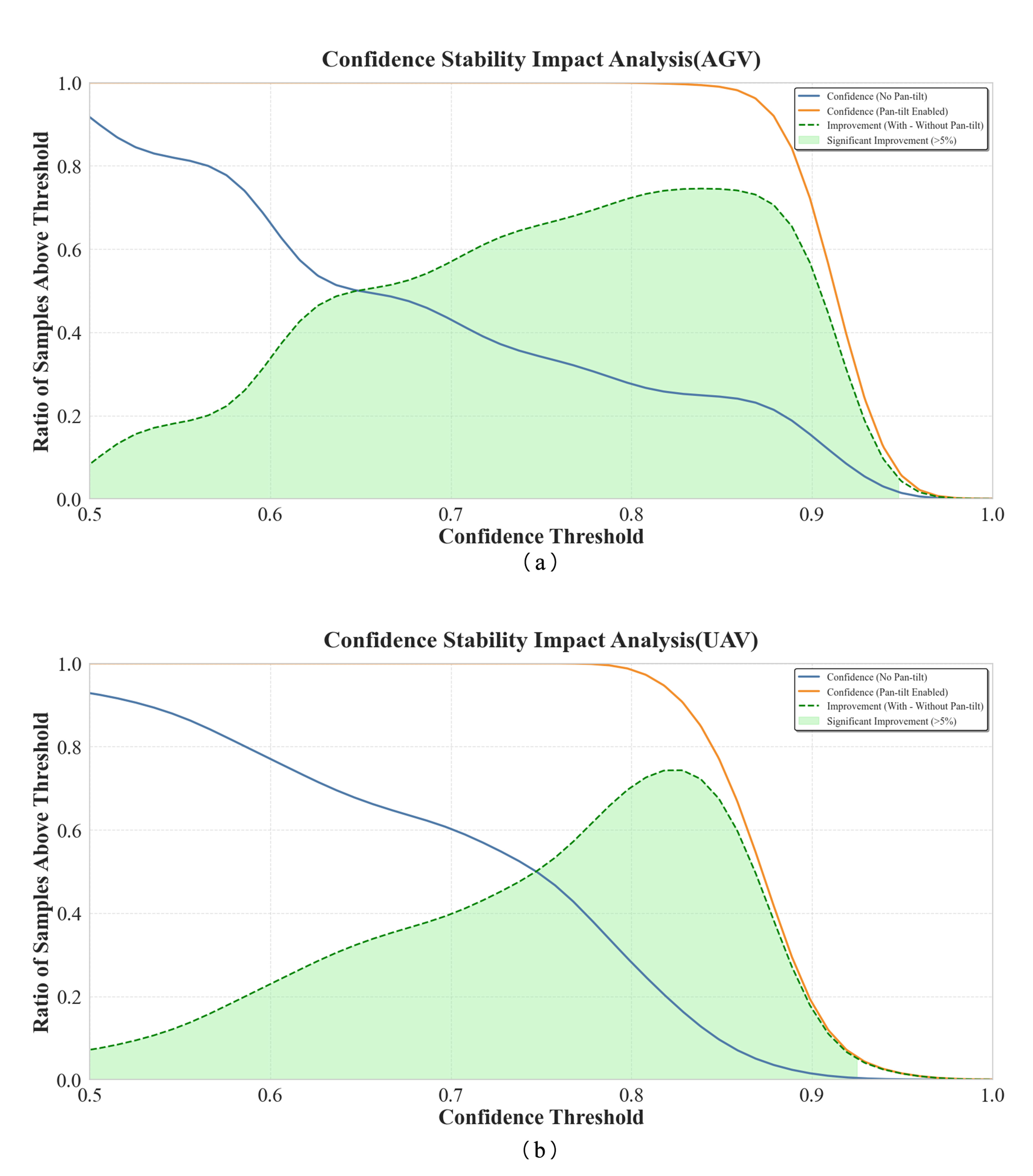}}
\caption{Comparison of confidence levels in AGV (a) and UAV (b) detection with and without the gimbal activated. The solid orange line represents the confidence level when the gimbal is enabled, while the solid blue line indicates the confidence level when the gimbal is disabled. The green dashed line denotes the difference between the two curves, and the green shaded area highlights the portion where the difference exceeds 5\%.}
\label{fig19}
\end{figure}

\subsection{Experiment of Bio-Pan-Tilt in Real world} 
In this experiment, the target was placed within the camera’s field of view, following the same starting point and route to evaluate the impact of the pan-tilt system on the target detection confidence, comparing scenarios with and without the pan-tilt system \cite{ref31}.
\begin{figure*}[t]
\centerline{\includegraphics[width=1.0\linewidth]{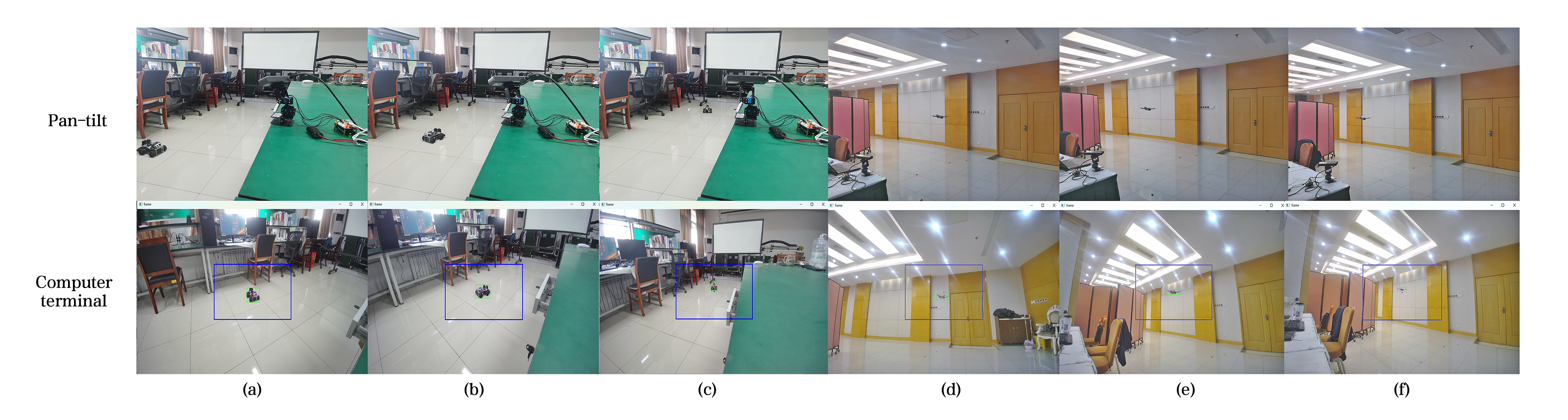}}
\caption{Bionic pan-tilt system tracking target experiment real picture, the upper part of the AGV tracking experiment, the lower part of the UAV tracking experiment, and target markers are displayed only when the confidence is greater than 0.5. Among them, the green box identifies the target, the red point identifies the target center, the blue point identifies the center of the screen, the upper side of the gimbal's positional state, the lower side of the real-time screen of the computer terminal}
\label{fig13}
\end{figure*}

\begin{figure}[t]
\centerline{\includegraphics[width=0.6\linewidth]{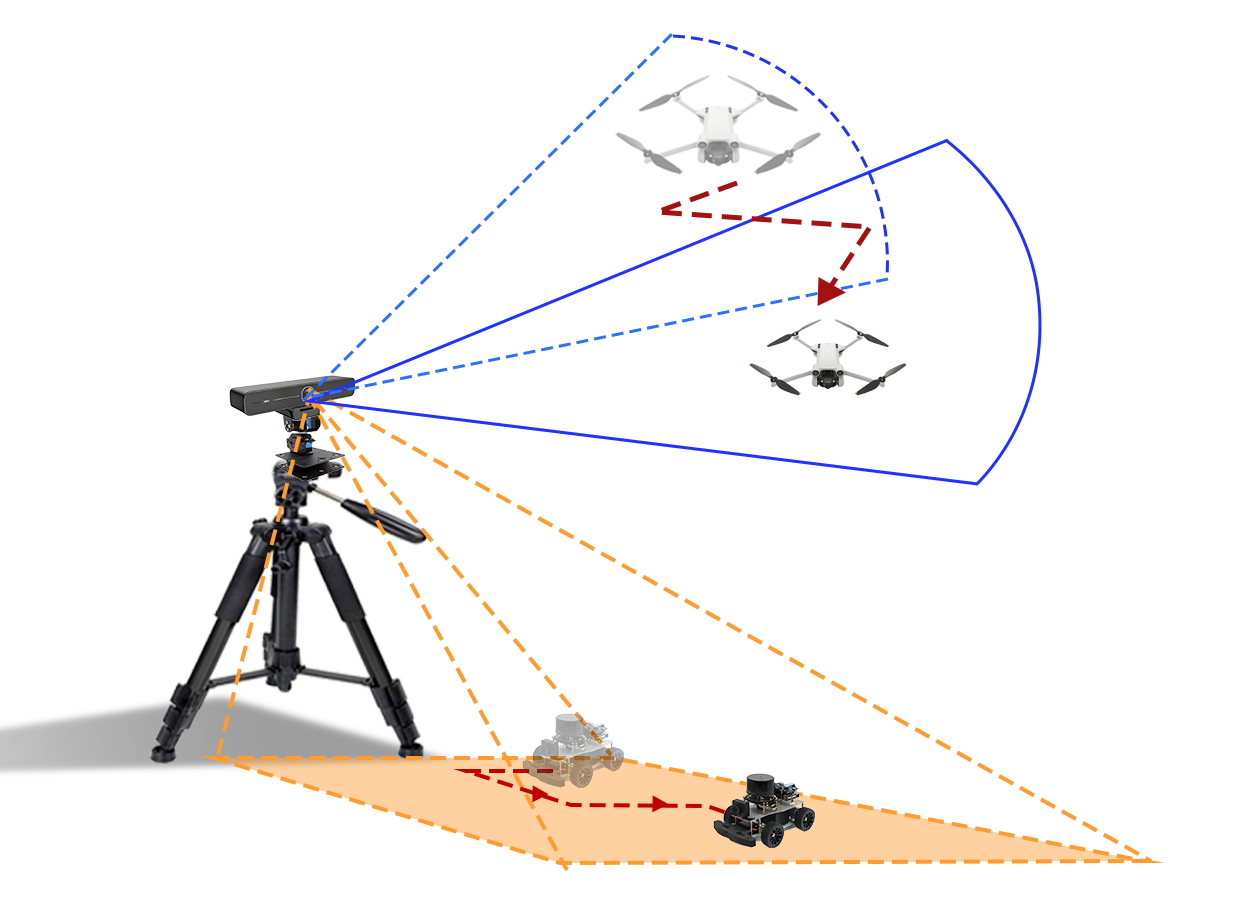}}
\caption{The camera on the tripod tracks the target vehicle moving along a predetermined path within its field of view, comparing the use and non use of the gimbal to evaluate its impact on the confidence of the detection algorithm}
\label{fig18}
\end{figure}

A standardized test environment was established to ensure that all conditions, such as lighting, starting point, and route, remained consistent, with the only variable being the use of the pan-tilt system. In the experiment, the target moved from the starting point to the endpoint along a predetermined trajectory, while the camera simultaneously captured real-time images of the vehicle throughout the movement. The experimental setup is shown in figure \ref{fig18}.

In the condition without a pan-tilt, the confidence evaluation results of the target detection algorithm on AGVs and UAVs were recorded. Subsequently, under the same test environment, the camera was mounted on the pan-tilt, and the corresponding confidence results were recorded again. To ensure the validity of the analysis, only samples with a confidence level greater than 0.5 were retained in the experiment, while detection results below this threshold were excluded due to insufficient reliability. The experimental results are shown in Figure \ref{fig19}.

The results reveal that the use of the pan-tilt significantly improves the stability of the target detection confidence. Without the pan-tilt, the confidence fluctuates significantly; however, with the pan-tilt, a higher proportion of samples are maintained under most confidence thresholds, particularly in the high-confidence region. This observation indicates that the assistance of the pan-tilt notably enhances the confidence and stability of target detection. While both data sets exhibit fluctuations, the dynamic view with pan-tilt assistance shows smaller fluctuations and maintains a more stable confidence proportion in the critical threshold interval. This further validates the essential role of the pan-tilt in improving target tracking and detection.

Figure \ref{fig13} presents a real-time image of the bionic pan-tilt system in operation. The experimental results demonstrate that the bionic pan-tilt system can effectively and stably track the target during its movement, keeping it within the central area of the frame. This confirms the effectiveness and feasibility of the proposed tracking system.

\section{Summary}
This study enhances the detection and tracking capabilities of small targets by optimizing the model framework and developing a bionic pan-tilt control strategy. CA-YOLO incorporates the MHSA module, small target detection head, and CFAM module to improve generalization performance and small target detection accuracy; the bionic pan-tilt integrates multiple strategies to enhance system stability and response speed. Experiments confirm that CA-YOLO performs prominently in small target detection accuracy, and the bionic pan-tilt shows significant potential in time-sensitive target tracking.

Despite the notable progress achieved in target detection and tracking in this study, there are still limitations. Firstly, the tracking system is based on a fixed pan-tilt, lacking a mobile carrier to support dynamic tracking. Secondly, target recognition relies on pre-trained models, making it difficult to handle new untrained targets. Future research will focus on addressing the system's limitations, developing dynamic carrier systems to enhance the flexibility of the pan-tilt, exploring universal target detection and tracking methods to improve adaptability to unknown targets, and providing support for fields such as UAV inspection and battlefield reconnaissance.

\bibliographystyle{unsrt} 
\bibliography{ref}

@article{ref1,
  title={Attention-Enhanced Co-Interactive Fusion Network (AECIF-Net) for automated structural condition assessment in visual inspection},
  author={ Zhang, Chenyu  and  Yin, Zhaozheng  and  Qin, Ruwen },
  journal={Automation in construction},
  number={Mar.},
  pages={159},
  year={2024},
}

@article{ref2,
  title={Improved YOLOv5 infrared tank target detection method under ground background.},
  author={ Liang, Chao  and  Yan, Z.  and  Ren, M.  and  Wu, Jiangpeng  and  Tian, L.  and  Guo, Xu  and  Li, Jie },
  journal={Scientific reports},
  volume={13 1},
  pages={
          6269
        },
  year={2023},
}

@article{ref3,
  title={YOLO-PowerLite: A Lightweight YOLO Model for Transmission Line Abnormal Target Detection},
  author={Chuanyao Liu and Shuangfeng Wei and Shaobo Zhong and Fan Yu},
  journal={IEEE Access},
  year={2024},
  volume={12},
  pages={105004-105015},
  url={https://api.semanticscholar.org/CorpusID:271557405}
}

@ARTICLE{ref4,
  author={Liu, Yunxiang and Luo, Peng},
  journal={IEEE Access}, 
  title={YOLO-TS: A Lightweight YOLO Model for Traffic Sign Detection}, 
  year={2024},
  volume={12},
  number={},
  pages={169013-169023},
  doi={10.1109/ACCESS.2024.3498057}}

@ARTICLE{ref5,
  author={Liu, Xingyu and Chu, Yuanfeng and Hu, Yiheng and Zhao, Nan},
  journal={IEEE Open Journal of Intelligent Transportation Systems}, 
  title={Enhancing Intelligent Road Target Monitoring: A Novel BGS-YOLO Approach Based on the YOLOv8 Algorithm}, 
  year={2024},
  volume={5},
  number={},
  pages={509-519},
  doi={10.1109/OJITS.2024.3449698}}

@ARTICLE{ref6,
  author={Xu, Qiwei and Lin, Runzi and Yue, Han and Huang, Hong and Yang, Yun and Yao, Zhigang},
  journal={IEEE Access}, 
  title={Research on Small Target Detection in Driving Scenarios Based on Improved Yolo Network}, 
  year={2020},
  volume={8},
  number={},
  pages={27574-27583},
  doi={10.1109/ACCESS.2020.2966328}}

@ARTICLE{ref7,
  author={Cheng, Liang},
  journal={IEEE Access}, 
  title={A Highly Robust Helmet Detection Algorithm Based on YOLO V8 and Transformer}, 
  year={2024},
  volume={12},
  number={},
  pages={130693-130705},
  doi={10.1109/ACCESS.2024.3459591}}

@ARTICLE{ref8,
  author={Zhang, Tinglin and Pang, Huanli and Jiang, Changhong},
  journal={IEEE Access}, 
  title={GDM-YOLO: A Model for Steel Surface Defect Detection Based on YOLOv8s}, 
  year={2024},
  volume={12},
  number={},
  pages={148817-148825},
  doi={10.1109/ACCESS.2024.3476908}}

@article{ref9,
  title={Road Surface Defect Detection Algorithm Based on YOLOv8},
  author={ Sun, Zhen  and  Zhu, Lingxi  and  Qin, Su  and  Yu, Yongbo  and  Ju, Ruiwen  and  Li, Qingdang },
  journal={Electronics (2079-9292)},
  volume={13},
  number={12},
  year={2024},
}

@ARTICLE{ref10,
  author={Wang, Hanjun and Luo, Shiyu and Wang, Qun},
  journal={IEEE Access}, 
  title={Improved YOLOv8n for Foreign-Object Detection in Power Transmission Lines}, 
  year={2024},
  volume={12},
  number={},
  pages={121433-121440},
  doi={10.1109/ACCESS.2024.3452782}}

@ARTICLE{ref11,
  author={Luo, Qiang and Wu, Chenbo and Wu, Guangjie and Li, Weiyi},
  journal={IEEE Access}, 
  title={A Small Target Strawberry Recognition Method Based on Improved YOLOv8n Model}, 
  year={2024},
  volume={12},
  number={},
  pages={14987-14995},
  doi={10.1109/ACCESS.2024.3356869}}

@article{ref12,
  title={A defect detection method for industrial aluminum sheet surface based on improved YOLOv8 algorithm},
  author={ Wang, Luyang  and  Zhang, Gongxue  and  Wang, Weijun  and  Chen, Jinyuan  and  Jiang, Xuyao  and  Yuan, Hai  and  Huang, Zucheng  and  Yang, Lei  and  Sun, Teng },
  journal={Frontiers in Physics},
  year={2024},
}

@article{ref13,
  title={Biological Eagle-Eye-Based Visual Platform for Target Detection},
  author={ Deng, Yimin  and  Duan, Haibin },
  journal={IEEE Transactions on Aerospace and Electronic Systems},
  year={2018},
}

@ARTICLE{ref14,
  author={Fu, Qiang and Wang, Shutai and Wang, Jin and Wu, Xiaoyang and He, Wei},
  journal={IEEE/ASME Transactions on Mechatronics}, 
  title={Target Detection and Localization for Flapping-Wing Robots Based on Biological Eagle-Eye Vision}, 
  year={2025},
  volume={30},
  number={2},
  pages={1590-1600},
  doi={10.1109/TMECH.2024.3424983}}

@INPROCEEDINGS{ref15,
  author={Wang, Shutai and Fu, Qiang and Hu, Yinhao and Zhang, Chunhua and He, Wei},
  booktitle={2021 China Automation Congress (CAC)}, 
  title={A Miniature Biological Eagle-Eye Vision System for Small Target Detection}, 
  year={2021},
  volume={},
  number={},
  pages={725-730},
  doi={10.1109/CAC53003.2021.9727530}}

@article{ref16,
  title={A Pan-tilt Camera Control System of UAV Visual Tracking Based on Biomimetic Eye},
  author={ Zou, Hairong  and  Gong, Zhenbang  and  Xie, Shaorong  and  Ding, Wei },
  journal={IEEE},
  year={2007},
}

@INPROCEEDINGS{ref17,
  author={Lin, Hong-Syuan and Ma, Ming-You and Wu, Yuan-Ting and Lin, Wei-Cheng and Shen, Shang-En and Huang, Yi-Cheng},
  booktitle={2024 International Conference on System Science and Engineering (ICSSE)}, 
  title={Remote Surveillance Object Detection Based on YOLO with Gimbal Camera Tracking Control}, 
  year={2024},
  volume={},
  number={},
  pages={1-6},
  doi={10.1109/ICSSE61472.2024.10609000}}

@INPROCEEDINGS{ref18,
  author={Lin, Chin E. and Yang, Sheng-Kai},
  booktitle={2014 CACS International Automatic Control Conference (CACS 2014)}, 
  title={Camera gimbal tracking from UAV flight control}, 
  year={2014},
  volume={},
  number={},
  pages={319-322},
  doi={10.1109/CACS.2014.7097209}}

@INPROCEEDINGS{ref19,
  author={Pătru, George-Cristian and Vasilescu, Iuliu and Rosner, Daniel and Tudose, Dan},
  booktitle={2021 23rd International Conference on Control Systems and Computer Science (CSCS)}, 
  title={Aerial Drone Platform for Asset Tracking Using an Active Gimbal}, 
  year={2021},
  volume={},
  number={},
  pages={138-142},
  doi={10.1109/CSCS52396.2021.00030}}

@INPROCEEDINGS{ref20,
  author={Whitacre, William and Campbell, Mark and Wheeler, Matthew and Stevenson, Davis},
  booktitle={2007 American Control Conference}, 
  title={Flight Results from Tracking Ground Targets Using SeaScan UAVs with Gimballing Cameras}, 
  year={2007},
  volume={},
  number={},
  pages={377-383},
  doi={10.1109/ACC.2007.4282696}}

@article{ref21,
title = {Study on a Kalman Filter based PID Controller},
journal = {IFAC-PapersOnLine},
volume = {51},
number = {4},
pages = {422-425},
year = {2018},
note = {3rd IFAC Conference on Advances in Proportional-Integral-Derivative Control PID 2018},
issn = {2405-8963},
doi = {https://doi.org/10.1016/j.ifacol.2018.06.131},
url = {https://www.sciencedirect.com/science/article/pii/S2405896318304282},
author = {Shin Wakitani and Hiroki Nakanishi and Yoichiro Ashida and Toru Yamamoto}
}

@article{ref22,
  title={A Novel Method of Motion People Detection and Tracking in Moving Background},
  author={ Huang, Lve  and  Yan, Hua Biao  and  Tan, Lu Min },
  journal={Applied Mechanics and Materials},
  volume={373-375},
  pages={547-551},
  year={2013},
}

@INPROCEEDINGS{ref23,
  author={Manecy, Augustin and Diperi, Julien and Boyron, Marc and Marchand, Nicolas and Viollet, Stéphane},
  booktitle={2016 IEEE International Conference on Robotics and Automation (ICRA)}, 
  title={A novel hyperacute gimbal eye to implement precise hovering and target tracking on a quadrotor}, 
  year={2016},
  volume={},
  number={},
  pages={3212-3218},
  doi={10.1109/ICRA.2016.7487490}}

@misc{ref24,
      title={SOD-YOLOv8 -- Enhancing YOLOv8 for Small Object Detection in Traffic Scenes}, 
      author={Boshra Khalili and Andrew W. Smyth},
      year={2024},
      eprint={2408.04786},
      archivePrefix={arXiv},
      primaryClass={cs.CV},
      url={https://arxiv.org/abs/2408.04786}, 
}

@INPROCEEDINGS{ref25,
  author={Dai, Xiyang and Chen, Yinpeng and Xiao, Bin and Chen, Dongdong and Liu, Mengchen and Yuan, Lu and Zhang, Lei},
  booktitle={2021 IEEE/CVF Conference on Computer Vision and Pattern Recognition (CVPR)}, 
  title={Dynamic Head: Unifying Object Detection Heads with Attentions}, 
  year={2021},
  volume={},
  number={},
  pages={7369-7378},
  doi={10.1109/CVPR46437.2021.00729}}

@INPROCEEDINGS{ref27,
  author={Hongkai Chen and Xiaoguang Zhao and Min Tan},
  booktitle={Proceeding of the 11th World Congress on Intelligent Control and Automation}, 
  title={A novel pan-tilt camera control approach for visual tracking}, 
  year={2014},
  volume={},
  number={},
  pages={2860-2865},
  doi={10.1109/WCICA.2014.7053182}}

@INPROCEEDINGS{ref31,
  author={Kumar, Prashant and Sonkar, Sarvesh and Ghosh, A.K and Philip, Deepu},
  booktitle={2020 7th International Conference on Control, Decision and Information Technologies (CoDIT)}, 
  title={Real-time vision-based tracking of a moving terrain target from Light Weight Fixed Wing UAV using gimbal control}, 
  year={2020},
  volume={1},
  number={},
  pages={154-159},
  doi={10.1109/CoDIT49905.2020.9263896}}

@ARTICLE{ref33,
  author={Han, Hong-Gui and Feng, Cheng-Cheng and Sun, Hao-Yuan and Qiao, Jun-Fei},
  journal={IEEE Transactions on Cybernetics}, 
  title={Dynamic Surface Intelligent Robust Control of Nonlinear Systems With Fixed-Time Sliding-Mode Observer}, 
  year={2024},
  volume={54},
  number={11},
  pages={6767-6779},
  doi={10.1109/TCYB.2024.3456089}}

@INPROCEEDINGS{ref34,
  author={Yosafat, S Robin and Machbub, Carmadi and Hidayat, Egi M. I.},
  booktitle={2017 7th IEEE International Conference on System Engineering and Technology (ICSET)}, 
  title={Design and implementation of Pan-Tilt control for face tracking}, 
  year={2017},
  volume={},
  number={},
  pages={217-222},
  doi={10.1109/ICSEngT.2017.8123449}}

@ARTICLE{ref37,
  author={Wang, Hai and Liu, Chenyu and Cai, Yingfeng and Chen, Long and Li, Yicheng},
  journal={IEEE Transactions on Instrumentation and Measurement}, 
  title={YOLOv8-QSD: An Improved Small Object Detection Algorithm for Autonomous Vehicles Based on YOLOv8}, 
  year={2024},
  volume={73},
  number={},
  pages={1-16},
  doi={10.1109/TIM.2024.3379090}}

@ARTICLE{ref38,
  author={Larsen, Martin Vonheim and Mathiassen, Kim},
  journal={IEEE Transactions on Robotics}, 
  title={Achieving Subpixel Platform Accuracy With Pan–Tilt–Zoom Cameras in Uncertain Times}, 
  year={2025},
  volume={41},
  number={},
  pages={628-646},
  doi={10.1109/TRO.2024.3508141}}

@incollection{ref39,
title = {Anatomy and Physiology},
editor = { J. Gordon Betts, Kelly A. Young, James A.},
booktitle = {Anatomy and Physiology},
publisher = {OpenStax},
year = {2023},
url = {https://openstax.org/books/anatomy-and-physiology/pages/1-introduction},
author = { J. Gordon Betts, Kelly A. Young, James A.}
}

@ARTICLE{ref40,
  author={Zhang, Liqian and Zhang, Qing and Zhao, Rui},
  journal={IEEE Transactions on Circuits and Systems for Video Technology}, 
  title={Progressive Dual-Attention Residual Network for Salient Object Detection}, 
  year={2022},
  volume={32},
  number={9},
  pages={5902-5915},
  doi={10.1109/TCSVT.2022.3164093}}

@INPROCEEDINGS{ref43,
  author={Zou, Hairong and Gong, Zhenbang and Xie, Shaorong and Ding, Wei},
  booktitle={2006 IEEE International Conference on Robotics and Biomimetics}, 
  title={A Pan-tilt Camera Control System of UAV Visual Tracking Based on Biomimetic Eye}, 
  year={2006},
  volume={},
  number={},
  pages={1477-1482},
  doi={10.1109/ROBIO.2006.340147}}

@ARTICLE{ref44,
  author={Javanmard Alitappeh, Reza and John, Akhil and Dias, Bernardo and van Opstal, A. John and Bernardino, Alexandre},
  journal={IEEE Transactions on Cognitive and Developmental Systems}, 
  title={Emergence of Human Oculomotor Behavior in a Cable-Driven Biomimetic Robotic Eye Using Optimal Control}, 
  year={2024},
  volume={16},
  number={4},
  pages={1546-1560},
  doi={10.1109/TCDS.2024.3376072}}

@article{ref45,
  title={Research on motion target detection based on infrared biomimetic compound eye camera},
  author={ Li, Linhan  and  Wang, Xiaoyu  and  Lei, Teng  and  Yue, Juan  and  Gao, Sili  and  Yu, Yang  and  Su, Haifeng },
  year={2024},
  journal={Scientific Reports},
}

@ARTICLE{ref46,
  author={Li, Fuzhang and Lin, Chuan and Zhang, Qing and Wang, Ruixing},
  journal={IEEE Access}, 
  title={A Biologically Inspired Contour Detection Model Based on Multiple Visual Channels and Multi-Hierarchical Visual Information}, 
  year={2020},
  volume={8},
  number={},
  pages={15410-15422},
  doi={10.1109/ACCESS.2020.2966916}}

@ARTICLE{ref47,
  author={Wang, Hongxin and Zhao, Jiannan and Wang, Huatian and Hu, Cheng and Peng, Jigen and Yue, Shigang},
  journal={IEEE Transactions on Cybernetics}, 
  title={Attention and Prediction-Guided Motion Detection for Low-Contrast Small Moving Targets}, 
  year={2023},
  volume={53},
  number={10},
  pages={6340-6352},
  doi={10.1109/TCYB.2022.3170699}}

@unknown{ref48,
author = {Ge, Zheng and Liu, Songtao and Wang, Feng and Li, Zeming and Sun, Jian},
year = {2021},
month = {07},
pages = {},
title = {YOLOX: Exceeding YOLO Series in 2021},
doi = {10.48550/arXiv.2107.08430}
}

@article{ref49,
  title={YOLOv9: Learning What You Want to Learn Using Programmable Gradient Information},
  author={Chien-Yao Wang and I-Hau Yeh and Hongpeng Liao},
  journal={ArXiv},
  year={2024},
  volume={abs/2402.13616},
  url={https://api.semanticscholar.org/CorpusID:267770251}
}

@article{ref50,
  title={YOLOv3: An Incremental Improvement},
  author={ Redmon, Joseph  and  Farhadi, Ali },
  journal={arXiv e-prints},
  year={2018},
}

@ARTICLE{ref51,
       author = {{Khanam}, Rahima and {Hussain}, Muhammad},
        title = "{YOLOv11: An Overview of the Key Architectural Enhancements}",
      journal = {arXiv e-prints},
         year = 2024,
        month = oct,
          eid = {arXiv:2410.17725},
        pages = {arXiv:2410.17725},
          doi = {10.48550/arXiv.2410.17725},
archivePrefix = {arXiv},
       eprint = {2410.17725},
 primaryClass = {cs.CV},
       adsurl = {https://ui.adsabs.harvard.edu/abs/2024arXiv241017725K}
}

@INPROCEEDINGS {ref52,
  author={Duan, Kaiwen and Bai, Song and Xie, Lingxi and Qi, Honggang and Huang, Qingming and Tian, Qi},
  booktitle={2019 IEEE/CVF International Conference on Computer Vision (ICCV)}, 
  title={CenterNet: Keypoint Triplets for Object Detection}, 
  year={2019},
  volume={},
  number={},
  pages={6568-6577},
  doi={10.1109/ICCV.2019.00667}}

@ARTICLE{ref53,
       author = {{Carion}, Nicolas and {Massa}, Francisco and {Synnaeve}, Gabriel and {Usunier}, Nicolas and {Kirillov}, Alexander and {Zagoruyko}, Sergey},
        title = "{End-to-End Object Detection with Transformers}",
      journal = {arXiv e-prints},
         year = 2020,
        month = may,
          eid = {arXiv:2005.12872},
        pages = {arXiv:2005.12872},
          doi = {10.48550/arXiv.2005.12872},
archivePrefix = {arXiv},
       eprint = {2005.12872},
 primaryClass = {cs.CV},
       adsurl = {https://ui.adsabs.harvard.edu/abs/2020arXiv200512872C}
}

@ARTICLE{ref54,
  author={Ren, Shaoqing and He, Kaiming and Girshick, Ross and Sun, Jian},
  journal={IEEE Transactions on Pattern Analysis and Machine Intelligence}, 
  title={Faster R-CNN: Towards Real-Time Object Detection with Region Proposal Networks}, 
  year={2017},
  volume={39},
  number={6},
  pages={1137-1149},
  doi={10.1109/TPAMI.2016.2577031}}

@INPROCEEDINGS {ref55,
  author={Zhu, Chenchen and He, Yihui and Savvides, Marios},
  booktitle={2019 IEEE/CVF Conference on Computer Vision and Pattern Recognition (CVPR)}, 
  title={Feature Selective Anchor-Free Module for Single-Shot Object Detection}, 
  year={2019},
  volume={},
  number={},
  pages={840-849},
  doi={10.1109/CVPR.2019.00093}}

@ARTICLE{ref56,
       author = {{He}, Kaiming and {Gkioxari}, Georgia and {Doll{\'a}r}, Piotr and {Girshick}, Ross},
        title = "{Mask R-CNN}",
      journal = {arXiv e-prints},
         year = 2017,
        month = mar,
          eid = {arXiv:1703.06870},
        pages = {arXiv:1703.06870},
          doi = {10.48550/arXiv.1703.06870},
archivePrefix = {arXiv},
       eprint = {1703.06870},
 primaryClass = {cs.CV},
       adsurl = {https://ui.adsabs.harvard.edu/abs/2017arXiv170306870H}
}

@INPROCEEDINGS {ref57,
  author={Lin, Tsung-Yi and Goyal, Priya and Girshick, Ross and He, Kaiming and Dollár, Piotr},
  booktitle={2017 IEEE International Conference on Computer Vision (ICCV)}, 
  title={Focal Loss for Dense Object Detection}, 
  year={2017},
  volume={},
  number={},
  pages={2999-3007},
  doi={10.1109/ICCV.2017.324}}

@ARTICLE{ref58,
       author = {{Liu}, Wei and {Anguelov}, Dragomir and {Erhan}, Dumitru and {Szegedy}, Christian and {Reed}, Scott and {Fu}, Cheng-Yang and {Berg}, Alexander C.},
        title = "{SSD: Single Shot MultiBox Detector}",
      journal = {arXiv e-prints},
         year = 2015,
        month = dec,
          eid = {arXiv:1512.02325},
        pages = {arXiv:1512.02325},
          doi = {10.48550/arXiv.1512.02325},
archivePrefix = {arXiv},
       eprint = {1512.02325},
 primaryClass = {cs.CV},
       adsurl = {https://ui.adsabs.harvard.edu/abs/2015arXiv151202325L}
}

@article{ref60,
title = {Small and dim target detection in infrared imagery: A review, current techniques and future directions},
journal = {Neurocomputing},
volume = {630},
pages = {129640},
year = {2025},
issn = {0925-2312},
doi = {https://doi.org/10.1016/j.neucom.2025.129640},
url = {https://www.sciencedirect.com/science/article/pii/S0925231225003121},
author = {Nikhil Kumar and Pravendra Singh}
}

@article{ref61,
  title={Microsoft COCO: Common Objects in Context},
  author={Lin, Tsung Yi  and  Maire, Michael  and  Belongie, Serge  and  Hays, James  and  Zitnick, C. Lawrence },
  journal={Springer International Publishing},
  year={2014},
}

@inproceedings{ref62,
  title={VisDrone: A Benchmark for Object Detection and Tracking in Aerial Images and Videos},
  author={Pengfei Zhu and Longyin Wen and Xiao Bian and Haibin Ling and Qinghua Hu},
  booktitle={Proceedings of the IEEE/CVF International Conference on Computer Vision (ICCV)},
  year={2019}
}

@ARTICLE{ref63,
  author={Chen, Zhaodong and Ji, Hongbing and Zhang, Yongquan and Zhu, Zhigang and Li, Yifan},
  journal={IEEE Transactions on Circuits and Systems for Video Technology}, 
  title={High-Resolution Feature Pyramid Network for Small Object Detection on Drone View}, 
  year={2024},
  volume={34},
  number={1},
  pages={475-489},
  doi={10.1109/TCSVT.2023.3286896}}

@ARTICLE{ref64,
  author={Zhu, Zhiqin and Zheng, Renzhong and Qi, Guanqiu and Li, Shuang and Li, Yuanyuan and Gao, Xinbo},
  journal={IEEE Transactions on Circuits and Systems for Video Technology}, 
  title={Small Object Detection Method Based on Global Multi-Level Perception and Dynamic Region Aggregation}, 
  year={2024},
  volume={34},
  number={10},
  pages={10011-10022},
  doi={10.1109/TCSVT.2024.3402097}}

@ARTICLE{ref65,
  author={Huang, Ying and Jiang, Qinghan and Qian, Ying},
  journal={IEEE Transactions on Circuits and Systems for Video Technology}, 
  title={A Novel Method for Video Moving Object Detection Using Improved Independent Component Analysis}, 
  year={2021},
  volume={31},
  number={6},
  pages={2217-2230},
  doi={10.1109/TCSVT.2020.3023175}}

@ARTICLE{ref66,
  author={Shi, Guolong and Shen, Xinyi and Xiao, Fuke and He, Yigang},
  journal={IEEE Internet of Things Journal}, 
  title={DANTD: A Deep Abnormal Network Traffic Detection Model for Security of Industrial Internet of Things Using High-Order Features}, 
  year={2023},
  volume={10},
  number={24},
  pages={21143-21153},
  doi={10.1109/JIOT.2023.3253777}}

\vspace{25pt}
\vspace{-44pt}
\begin{IEEEbiography}[{\includegraphics[width=1in,height=1.25in,clip,keepaspectratio]{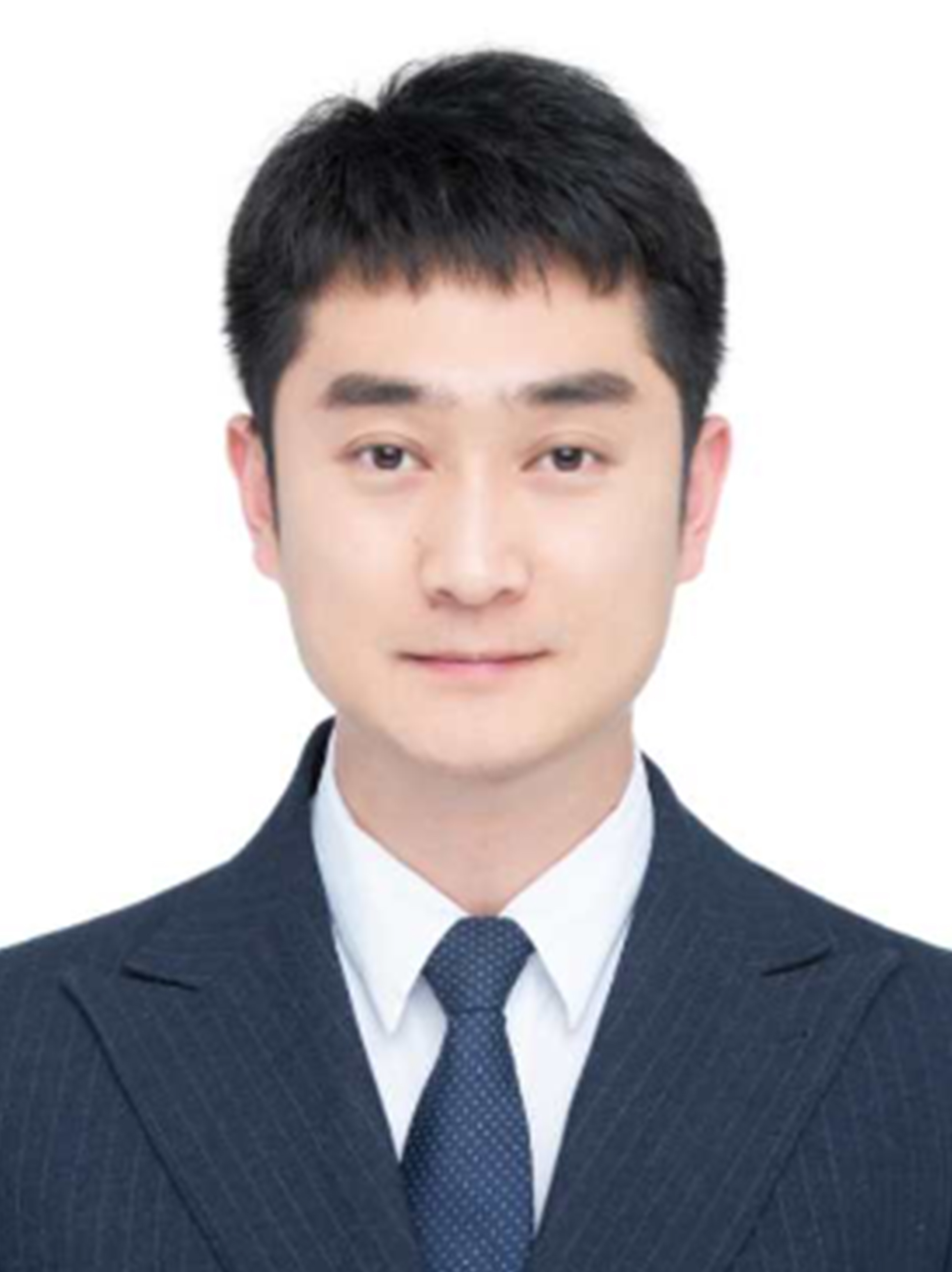}}]{Zhen Zhang} received his B.E. and M.E. degree in Mechanical Engineering from Xinjiang Agricultural University, China in 2012 and 2015, respectively, and the Ph.D. degree in Electrical Engineering and Computer Science from the Iwate University, Japan, in 2019. He is now working at the School of Mechanical And Automotive Engineering, Anhui Polytechnic University, as an associate peofessor. His current research interests include artificial intelligence, human-computer nature interaction system and robot intelligent control system. 
\end{IEEEbiography}

\begin{IEEEbiography}[{\includegraphics[width=1in,height=1.25in,clip,keepaspectratio]{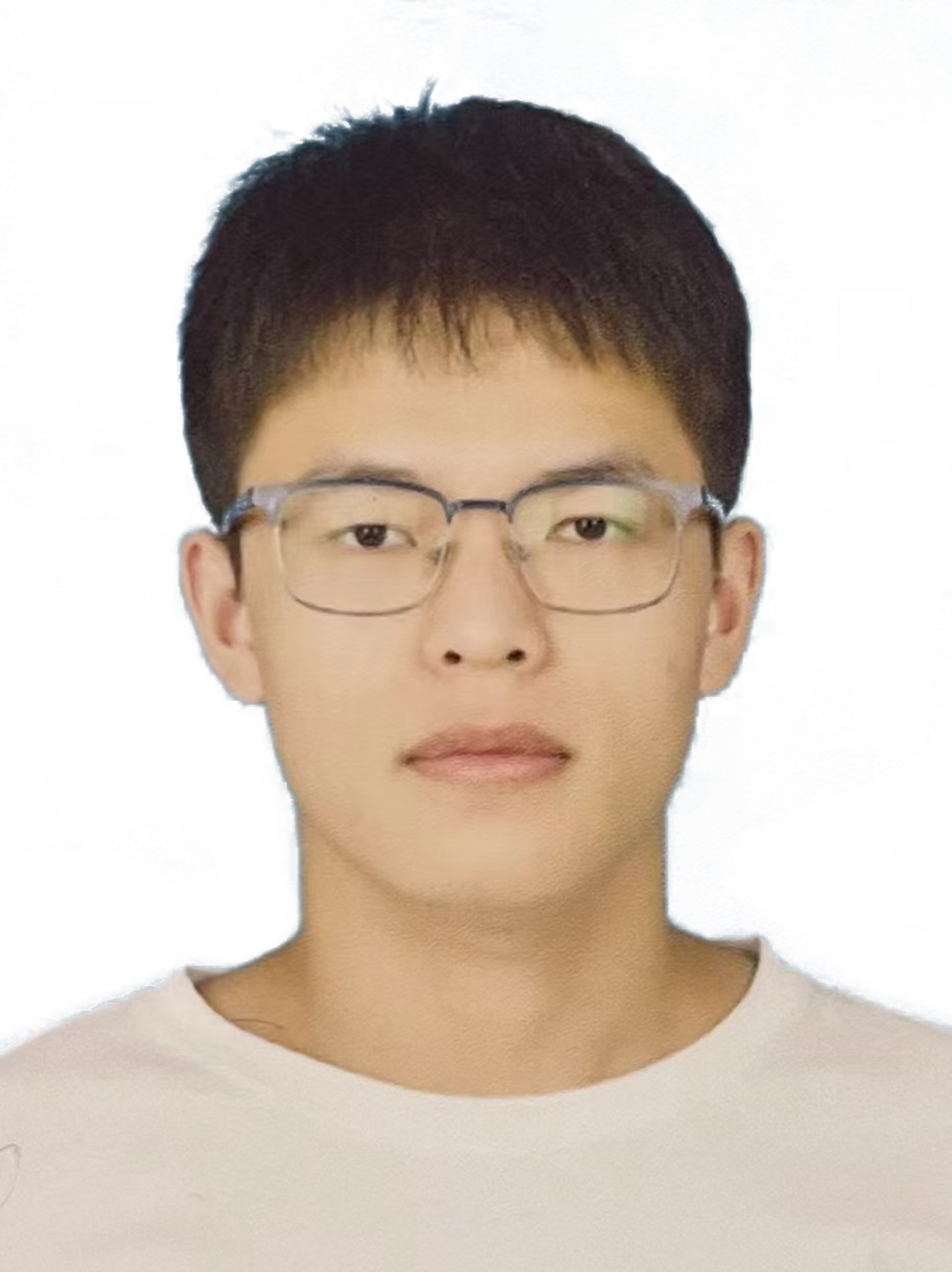}}]{Qing Zhao} received his B.E. degree in Mechanical Design, Manufacturing, and Automation from Anhui Polytechnic University, Wuhu, China, in 2023. He is currently pursuing a Master’s degree in the joint program between Anhui Polytechnic University and the National University of Defense Technology. His research interests include image processing and artificial intelligence. 
\end{IEEEbiography}

\begin{IEEEbiography}[{\includegraphics[width=1in,height=1.25in,clip,keepaspectratio]{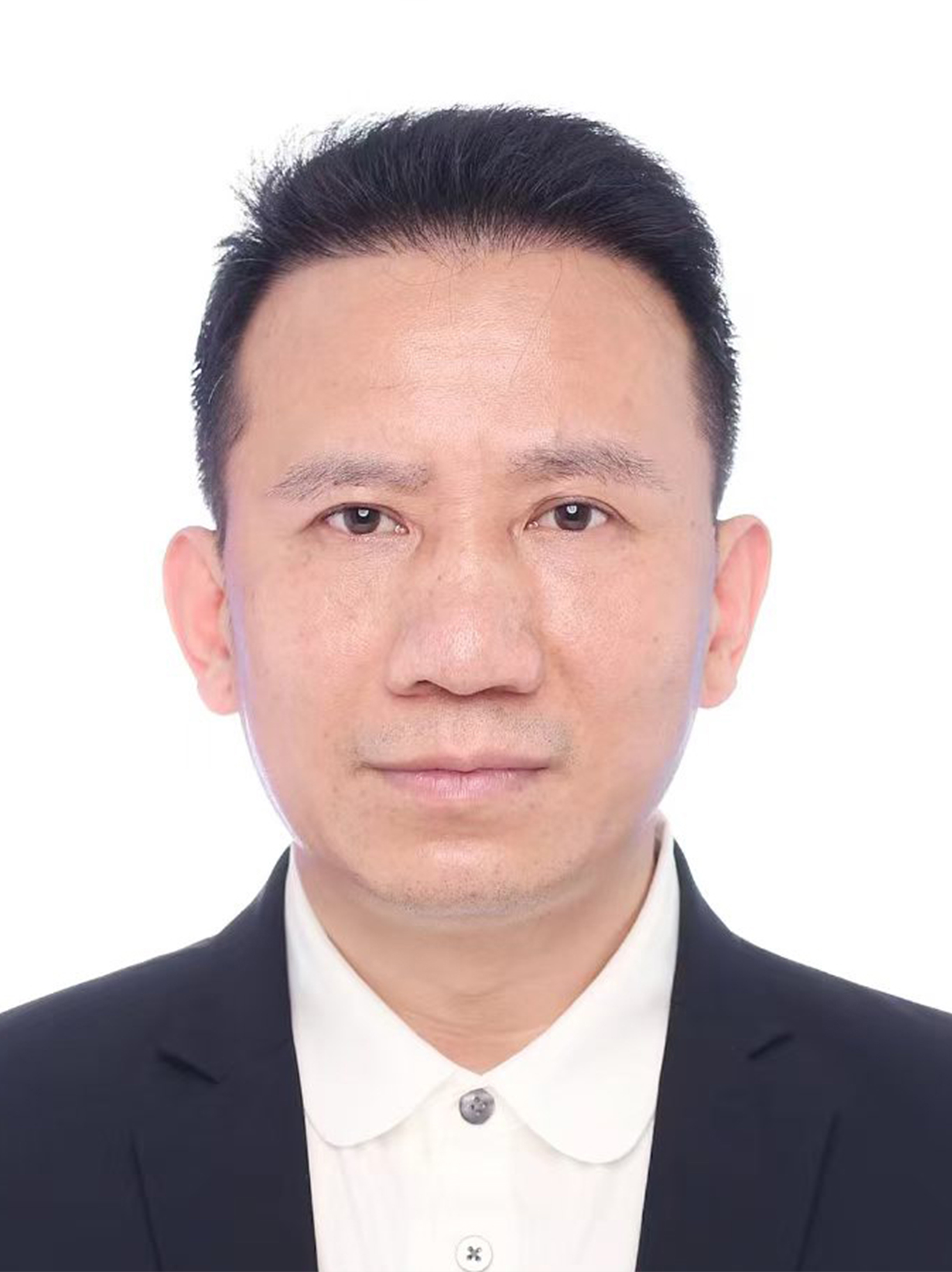}}]{Xiuhe Li} received his Ph.D. degree from the National University of Defense Technology, in 2004. He is now a professor in National University of Defense Technology. His research interests are radar signal processing and artificial intelligence.
\end{IEEEbiography}

\begin{IEEEbiography}[{\includegraphics[width=1in,height=1.25in,clip,keepaspectratio]{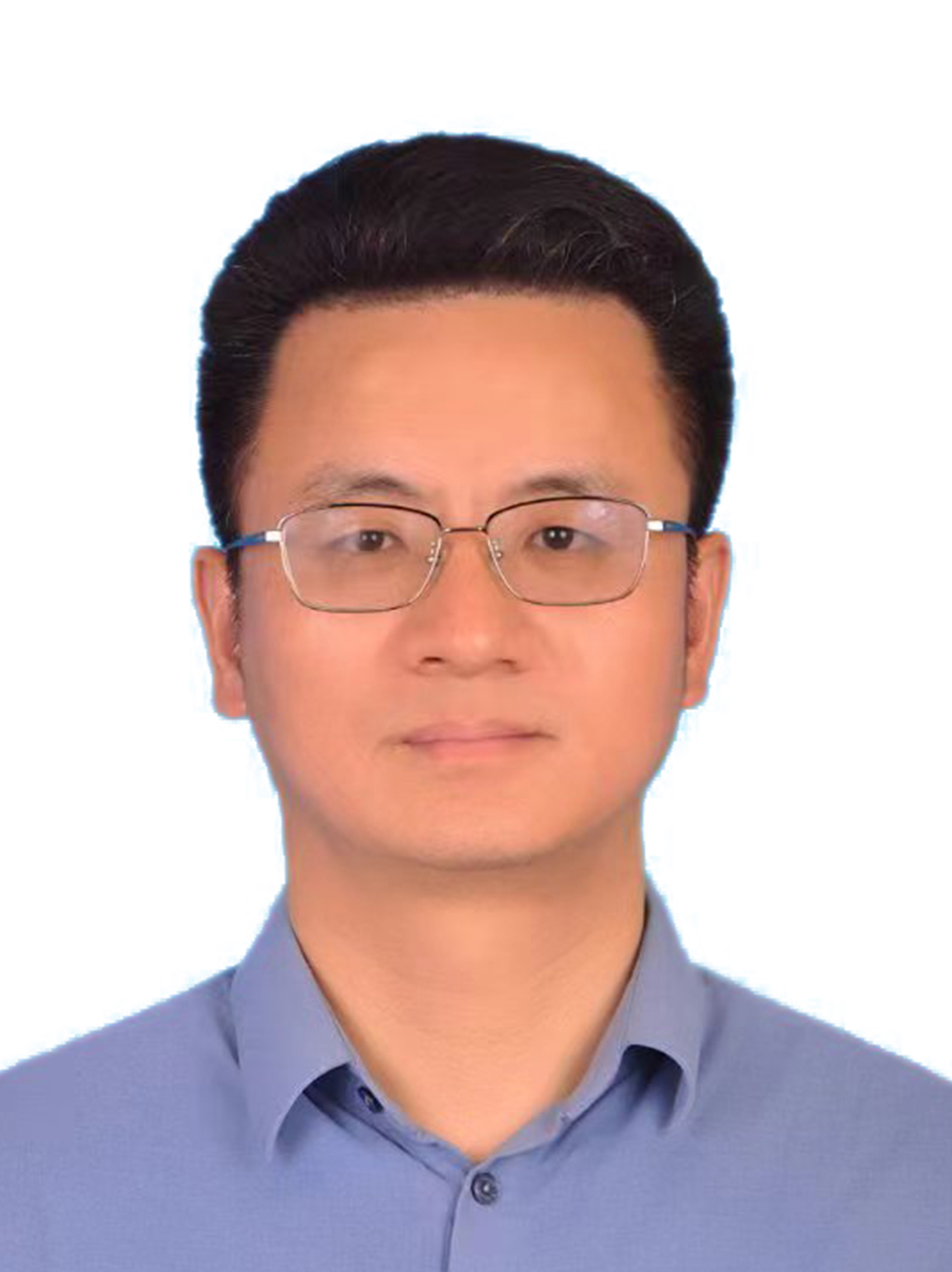}}]{Cheng Wang} received his master's degree from the College Electronic Engineering, in 2000. He is now a professor in National University of Defense Technology. His research interests are radar signal processing and artificial intelligence.
\end{IEEEbiography}

\begin{IEEEbiography}[{\includegraphics[width=1in,height=1.25in,clip,keepaspectratio]{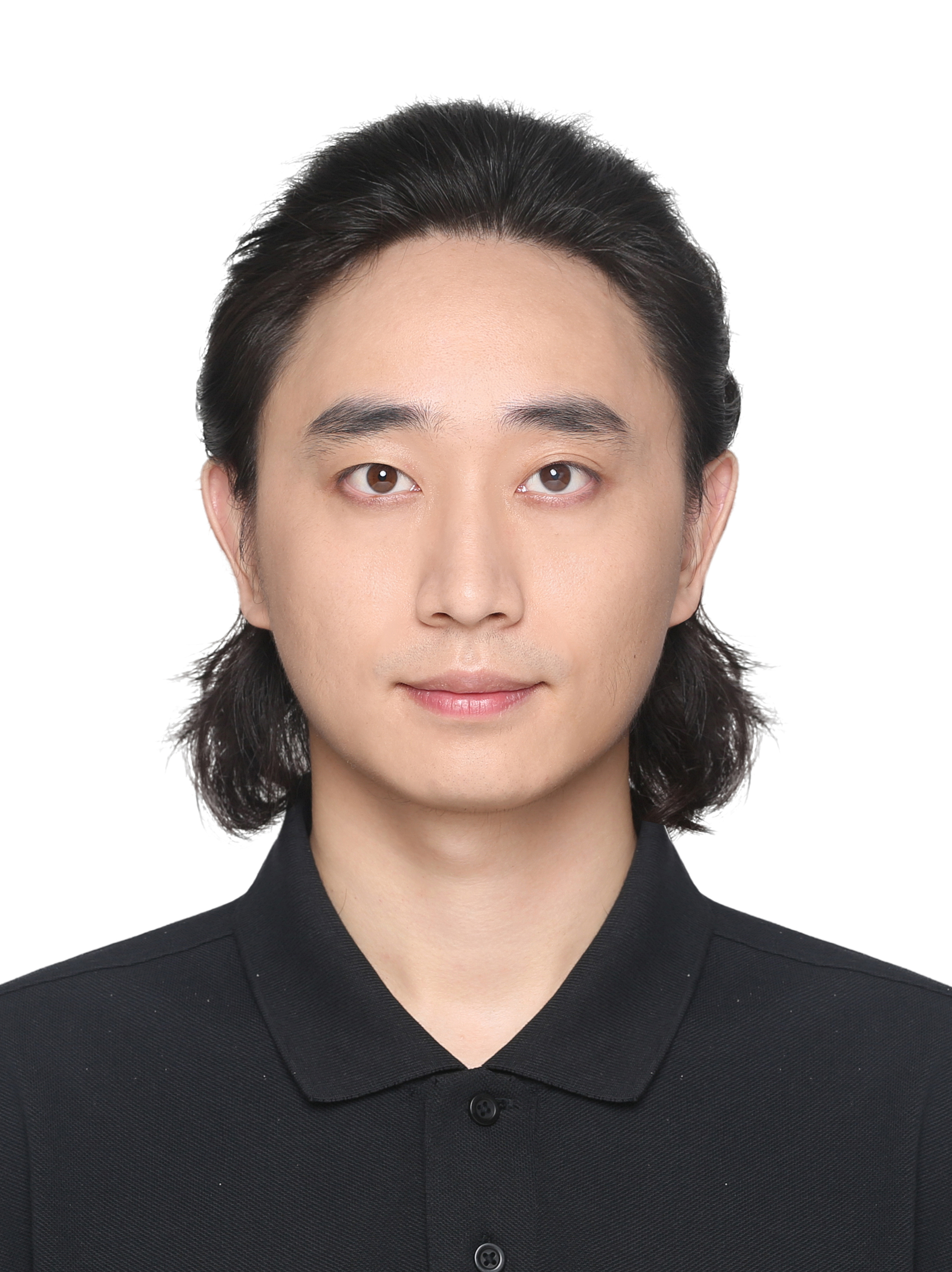}}]{Guoqiang Zhu} began his graduate studies at Tianjin University of Technology in 2020 and received his master’s degree in March 2023. He is currently working at the National University of Defense Technology, with primary research interests in deep learning and pattern recognition.
\end{IEEEbiography}

\begin{IEEEbiography}[{\includegraphics[width=1in,height=1.25in,clip,keepaspectratio]{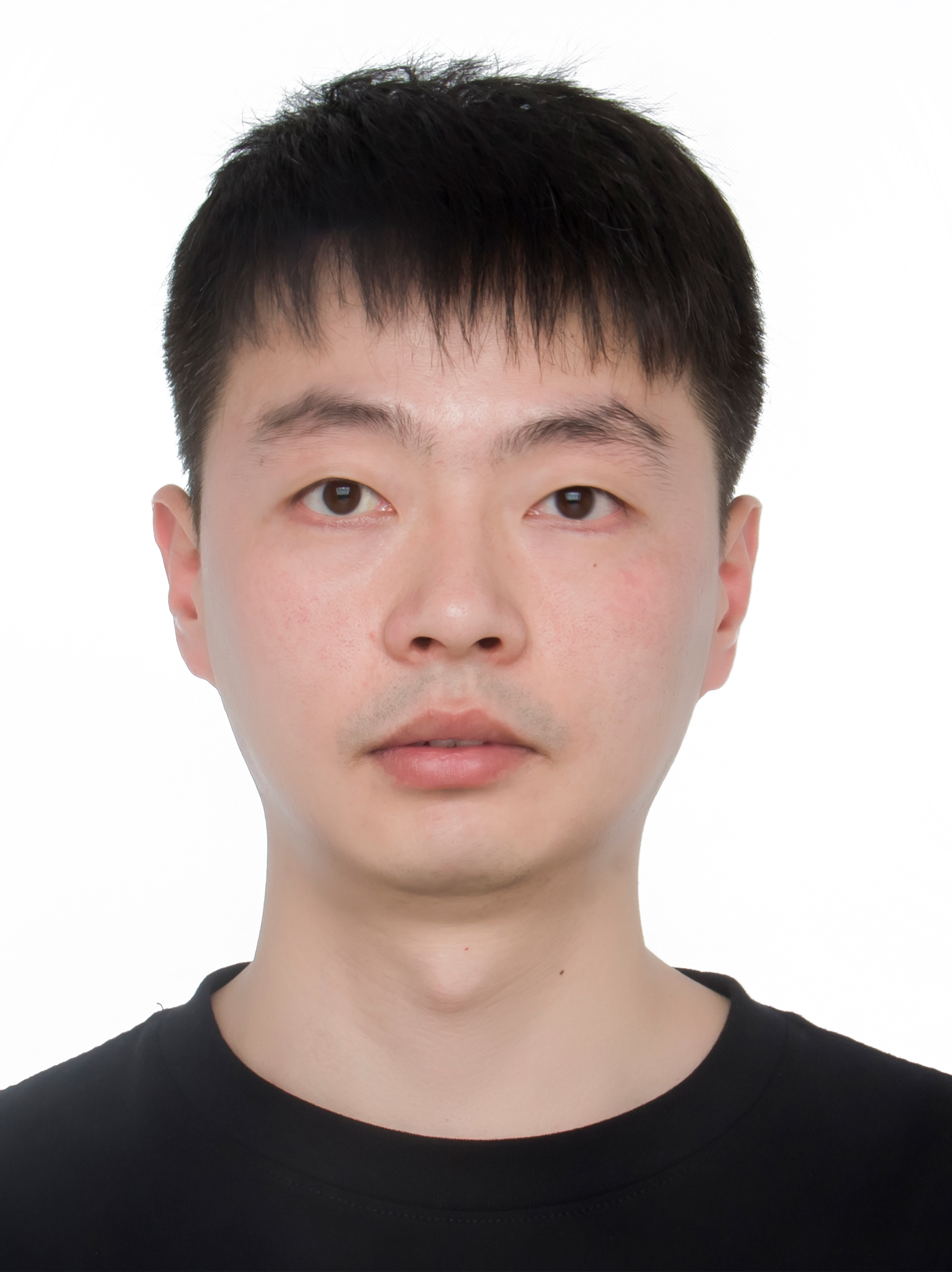}}]{Yu Zhang} received his B.E. in Computer Science and Technology from Northwest A\&F University in 2015 and his Master’s degree from Iwate University, Iwate, Japan, in 2018. He is currently working at the National University of Defense Technology in Hefei. His research interests include network intrusion detection, deep learning, and pattern recognition.
\end{IEEEbiography}

\begin{IEEEbiography}[{\includegraphics[width=1in,height=1.25in,clip,keepaspectratio]{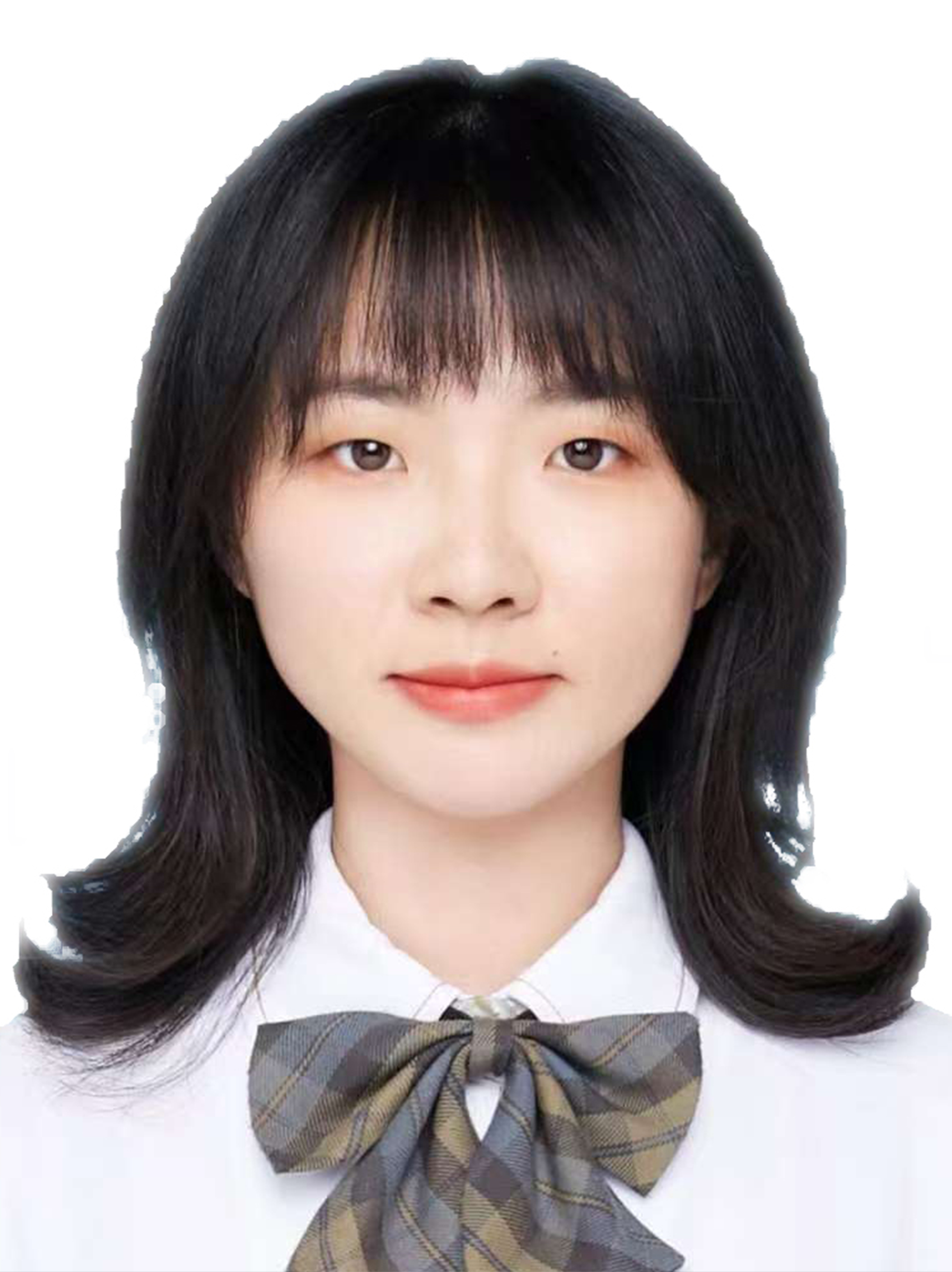}}]{Yining Huo} received the Master degree from the School of Advanced Technology, Xi'an Jiaotong-Liverpool University, Suzhou, China. She is currently working in National University of Defense Technology, Hefei. Her research interests include deep learning  and pattern recognition.
\end{IEEEbiography}

\begin{IEEEbiography}[{\includegraphics[width=1in,height=1.25in,clip,keepaspectratio]{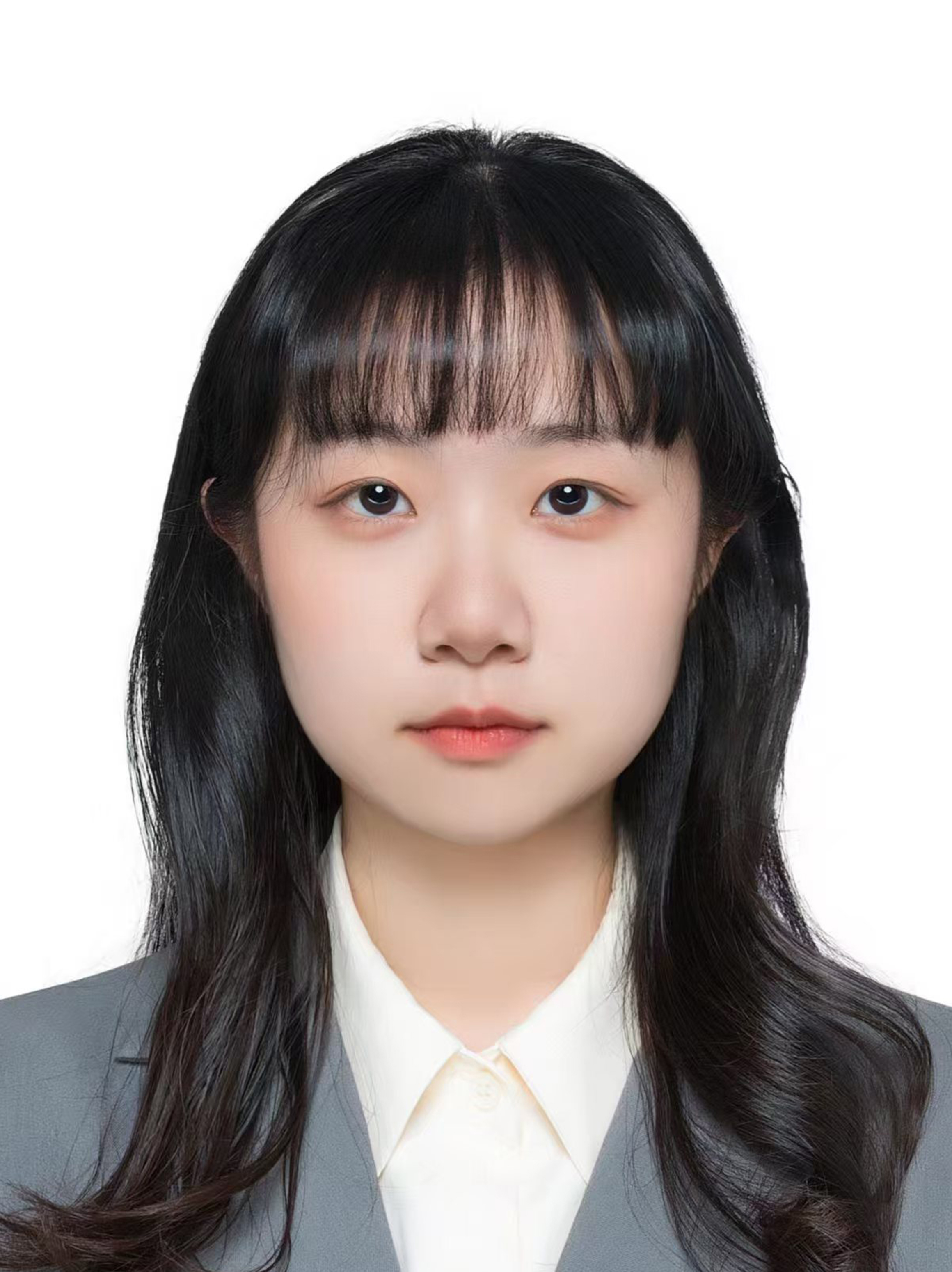}}]{Hongyi Yu} received her Master’s degree from Anhui Jianzhu University. She is currently working at the National University of Defense Technology, with art design and situational interface as her main research directions 
\end{IEEEbiography}

\begin{IEEEbiography}[{\includegraphics[width=1in,height=1.25in,clip,keepaspectratio]{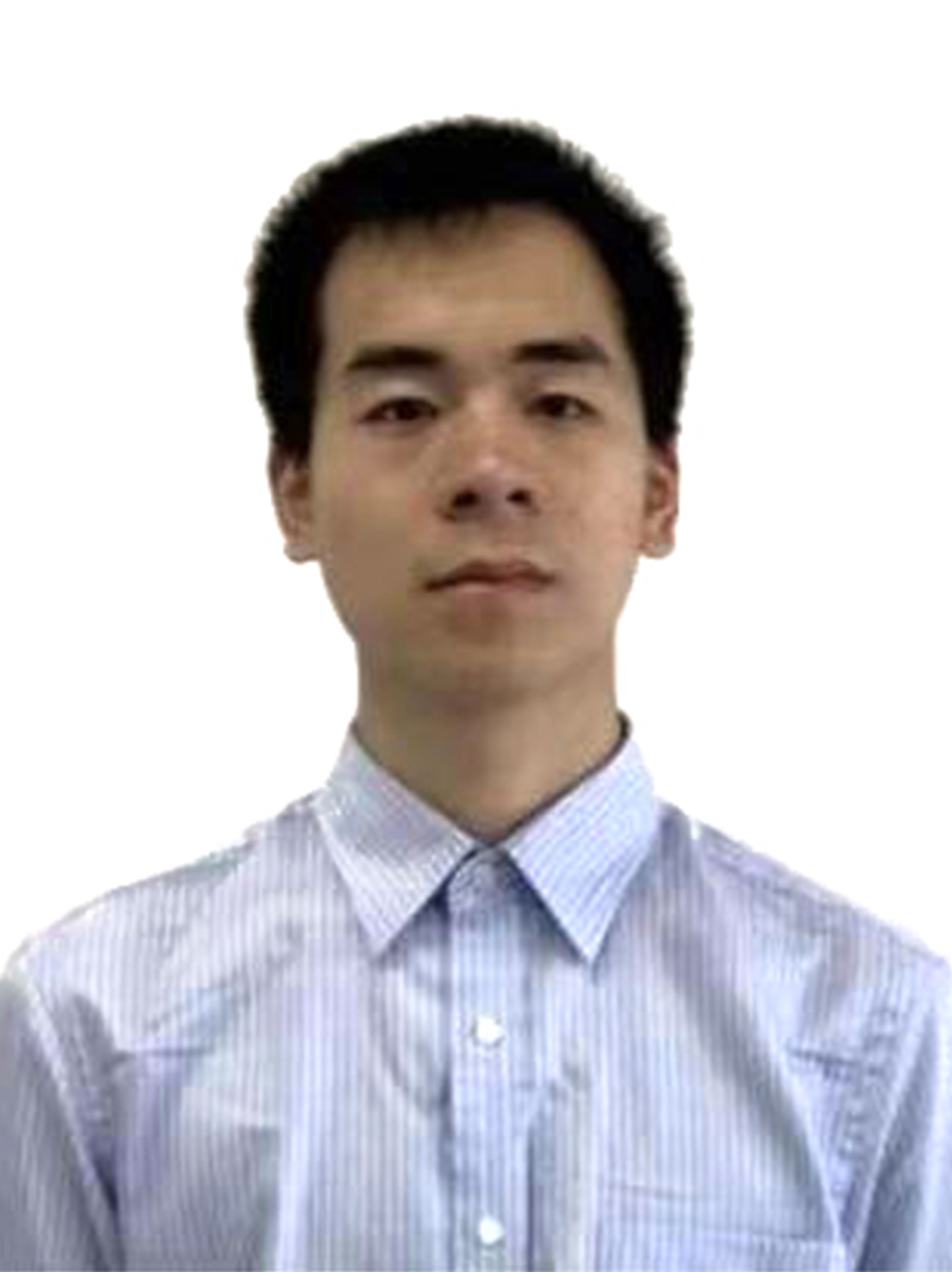}}]{Yi Zhang} received his B.E. in Computer Science and Technology from Northwest A\&F University in 2015, and his master’s and Ph.D. degrees from Iwate University in 2018 and 2021, respectively. He currently serves as an associate professor at the National University of Defense Technology. His research interests include artificial intelligence, foundation models, and image processing.
\end{IEEEbiography}

\vfill

\end{document}